\documentclass[a4paper,fleqn]{cas-dc}
\usepackage[square, numbers]{natbib}

\usepackage{caption}
\usepackage{cas-common}
\usepackage{graphicx}
\usepackage{diagbox}
\usepackage{mathtools}
\usepackage{amsmath}
\usepackage{bbding}
\usepackage{pifont}
\usepackage{wasysym}
\usepackage{tikz}
\usepackage{calc}
\usepackage{multirow}
\usepackage{rotating}
\usepackage{colortbl}
\usepackage{color}
\usepackage{ragged2e}
\usepackage{float}

\begin{document}
\captionsetup[figure]{labelfont={bf},labelformat={default},name={Fig.}, labelsep=period,}
\captionsetup[table]{labelfont={bf},labelformat={default},labelsep=newline,name={Table}, singlelinecheck=false}
\let\WriteBookmarks\relax
\def\floatpagepagefraction{1}
\def\textpagefraction{.001}

\shorttitle{Deep Learning Based 3D Segmentation: A Survey}    

\shortauthors{Y. He et~al. }  

\title [mode = title]{Deep Learning Based 3D Segmentation: A Survey}  



%


\author[1]{Yong He}[type=editor, auid=000, bioid=1, orcid= 0000-0003-2916-3068]

\ead{h.yong@hnu.edu.cn}

\author[1]{Hongshan Yu}[type=editor,auid=000,bioid=1, orcid = 0000-0003-1973-6766]
\cormark[1]
\ead{yuhongshancn@hotmail.com}
\cortext[1]{Corresponding author}

\author[1]{Xiaoyan Liu}[type=editor,auid=000,bioid=1,]
\ead{xiaoyan.liu@hnu.edu.cn}

\author[1]{Zhengeng Yang}[type=editor,auid=000,bioid=1,]
\ead{yzg050215@163.com}

\author[1]{Wei Sun}[type=editor,auid=000,bioid=1,]
\ead{david-sun@126.com}

\author[2]{Saeed Anwar}[type=editor,auid=000,bioid=1,]
\ead{saeed.anwar@kfupm.edu.sa}

\author[3]{Ajmal Mian}[type=editor,auid=000,bioid=1,]
\ead{ajmal.mian@uwa.edu.au}

\affiliation[1]{organization={Hunan University},
            addressline={Lushan South Rd., Yuelu Dist.}, 
            city={Changsha},
            postcode={410082}, 
            state={Hunan},
            country={China}}

\affiliation[2]{organization={The Australian National University},
            addressline={Acton}, 
            city={Canberra},
            postcode={2600}, 
            state={ACT},
            country={Australia}}

\affiliation[3]{organization={University of Western Australia},
            addressline={35 Stirling Hwy}, 
            city={Perth},
            postcode={6009}, 
            state={WA},
            country={Australia}}

\begin{abstract}
3D segmentation is a fundamental and challenging problem in computer vision with applications in autonomous driving and robotics. It has received significant attention from the computer vision, graphics and machine learning communities. Conventional methods for 3D segmentation, based on hand-crafted features and machine learning classifiers, lack generalization ability. Driven by their success in 2D computer vision, deep learning techniques have recently become the tool of choice for 3D segmentation tasks. This has led to an influx of many methods in the literature that have been evaluated on different benchmark datasets. Whereas survey papers on RGB-D and point cloud segmentation exist, there is a lack of a recent in-depth survey that covers all 3D data modalities and application domains. This paper fills the gap and comprehensively surveys the recent progress in deep learning-based 3D segmentation techniques. We cover over \textcolor{black}{220} works from the last six years, analyze their strengths and limitations, and discuss their competitive results on benchmark datasets. The survey provides a summary of the most commonly used pipelines and finally highlights promising research directions for the future.
\end{abstract}




\maketitle

\section{Introduction}

\begin{figure}[t]
\centering
\vspace{2mm}
\includegraphics[width=0.95\columnwidth]{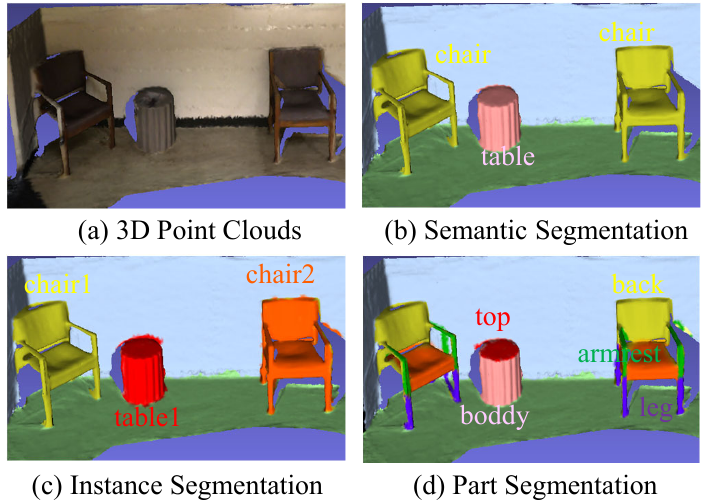}
\caption{The main three types of 3D segmentation are (b) 3D semantic segmentation, (c) 3D instance segmentation, and (d) 3D part segmentation on (a) 3D point clouds.}
\vspace{-2mm}
\label{dataandsegmentation}
\end{figure}

Segmentation of 3D scenes is a fundamental and challenging problem in computer vision as well as computer graphics. The objective of 3D segmentation is to build computational techniques that predict the fine-grained labels of objects in a 3D scene for a wide range of applications,
such as autonomous driving, mobile robots, industrial control, and augmented and virtual reality. As illustrated in Figure~\ref{dataandsegmentation}, 3D segmentation can be divided into semantic, instance and part segmentation. \textcolor{black}{Semantic segmentation aims to predict object class labels such as tables and chairs. Instance segmentation additionally distinguishes between different instances of the same class labels, e.g., chair one and two. Part segmentation aims to further decompose instances into various components, such as the same chair's armrests, legs, and backrests.} Semantic, instance, and part segmentation tasks show a progressive relationship in terms of semantic levels. Despite the difference in specific goals of three segmentation tasks, these tasks share a broader, unifying goal: they aim to segment 3D data into semantically meaningful regions based on certain criteria-whether it's by class,instance, part.

Compared to conventional single view 2D segmentation, 3D segmentation gives a more comprehensive understanding of a scene since  3D data (e.g., RGB-D, point cloud, voxel, mesh, 3D video) contain richer geometric, shape, and scale information with less background noise. Moreover, the representation of 3D data, for example, in the form of projected images, has more semantic information.

\begin{figure*}[t]
\centering
\vspace{3mm}
\includegraphics[width=0.95 \textwidth]{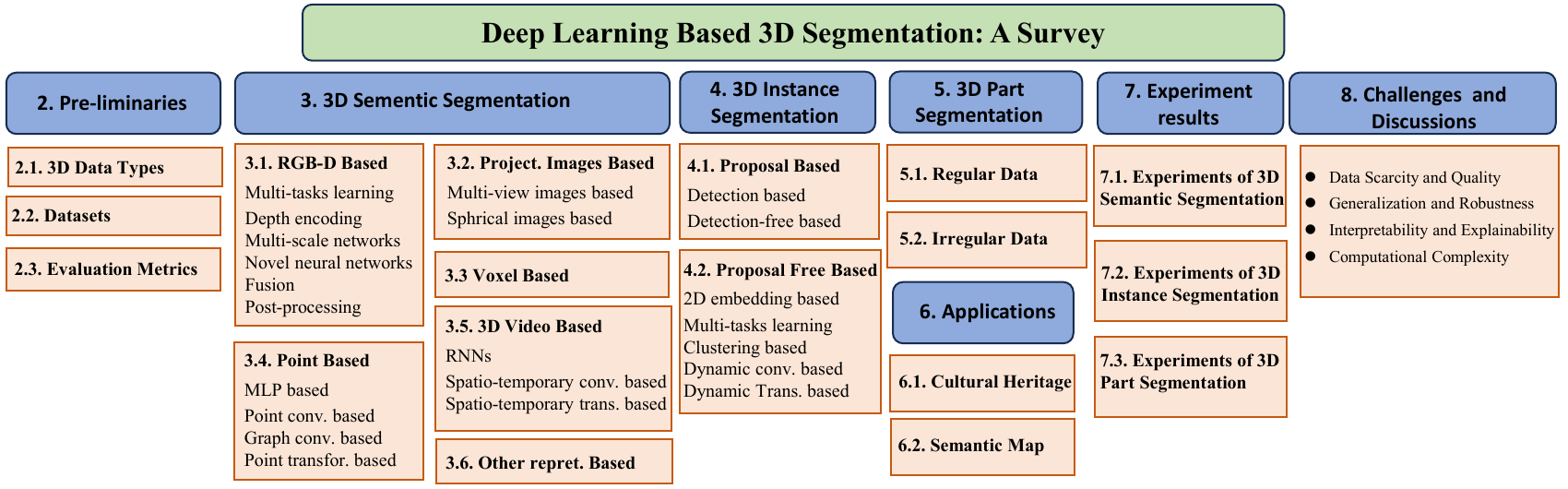}
\vspace{2mm}
\caption{Complete overview of the survey paper. Highlighting each section and its essential content.}

\label{overview}
\end{figure*}

Recently, deep learning techniques have dominated many research areas, including computer vision and natural language processing. Motivated by its success in learning powerful features, deep learning for 3D segmentation has also attracted growing interest from the research community over the past decade. However, 3D deep learning methods still face many unsolved challenges. For example, the irregularity of point clouds makes it difficult to exploit local features, and converting them to high-resolution voxels has a substantial computational burden.

This paper comprehensively surveys recent progress in deep learning methods for 3D segmentation. It focuses on analyzing commonly used building blocks, convolutional kernels and complete architectures, pointing out the pros and cons in each case. The survey covers over 180 representative articles published in the last six years. Although some notable 3D segmentation surveys have been released including RGB-D semantic segmentation~\cite{fooladgar2020survey}, remote sensing imagery segmentation~\cite{yuan2021review}, point clouds segmentation~\cite{xie2020linking},~\cite{guo2020deep},~\cite{liu2019deep},~\cite{bello2020deep},~\cite{naseer2018indoor},~\cite{ioannidou2017deep}, these surveys do not comprehensively cover all 3D data types and typical application domains. Most importantly, these surveys do not focus on 3D segmentation but give a general survey of deep learning from point clouds~\cite{guo2020deep},~\cite{liu2019deep},~\cite{bello2020deep},~\cite{naseer2018indoor},~\cite{ioannidou2017deep}.
Given the differences in domain knowledge required for semantic, instance, and part segmentation tasks in 3D segmentation, this paper reviews the deep learning techniques for each of these three segmentation tasks separately.
Figure~\ref{overview} shows the overview of the whole article. The contributions of this paper are summarized as follows:
\begin{itemize}

\item  To the best of our knowledge, this is the first survey paper to comprehensively cover deep learning methods on 3D segmentation \textcolor{black}{in computer vision}, covering the most common 3D data representations, including RGB-D, projected images, voxels, point clouds, meshes, and 3D videos.

\item This survey provides an in-depth analysis of the relative advantages and disadvantages of different 3D data segmentation methods. Unlike existing reviews, this survey focuses on deep learning methods designed specifically for 3D segmentation and discusses typical segmentation pipelines.

\item Finally, this survey provides comprehensive comparisons of existing methods on several public benchmark 3D datasets, draws interesting conclusions, and identifies promising future research directions.

\end{itemize}


\section{Pre-liminaries}
\label{section2}
This section introduces 3D data representations, popular 3D segmentation datasets, and evaluation metrics to help the reader easily navigate the rest of the survey.

\subsection{3D Data Types}

\begin{figure}[t]
\centering
\includegraphics[width=0.9\columnwidth]{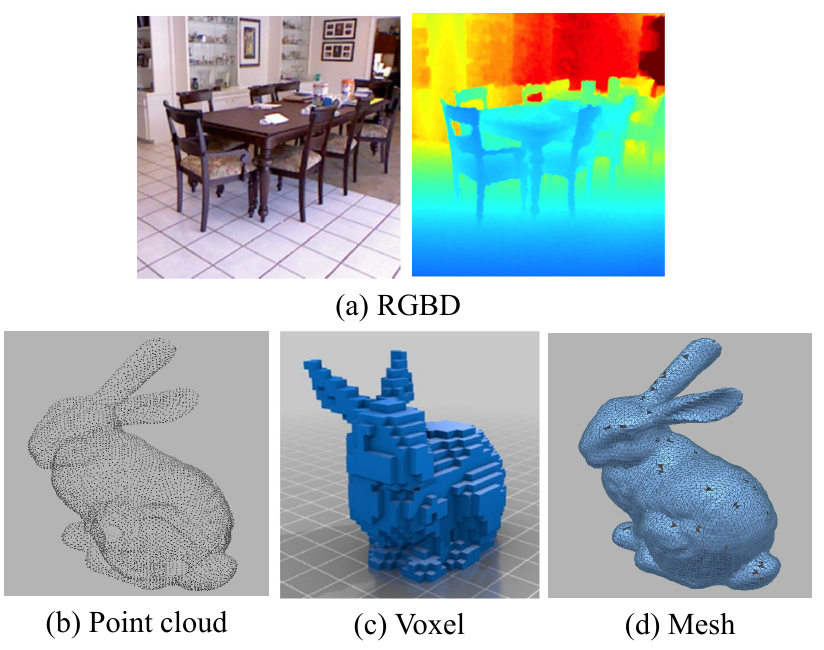}
\caption{The four types of 3D data are (a) RGBD, (b) Point cloud, (c) Voxel, and (d) Mesh.}
\label{data}
\end{figure}

With the rapid development of 3D sensors, different type of 3D data could be accessed easily.  While raw data originate from different sensors they might be transformed into a common 3D data format. For instance, raw 3D data generated from depth camera, terrestrial LiDAR, and mobile LiDAR can be transformed into a common point cloud format that includes coordinates and other attributes (e.g. RGB and surface normal).  We provide the details of the most common types of 3D data used in 3D segmentation, \textcolor{black}{specifically
for computer vision and 3D vision community}, which are visually shown in Figure~\ref{data}.

\vspace{2mm}\noindent\textbf{RGBD}: is a type of 3D data that combines RGB (Red, Green, Blue) color information with depth (D) information. It is commonly represented as a pair of images: a standard RGB image capturing color information and a depth image capturing distance information from the camera to objects in the scene. RGB-D data is typically obtained using depth-sensing devices like Microsoft Kinect, Intel RealSense, or structured light cameras. These devices emit infrared light or use other depth-sensing techniques to measure the distance to objects in the scene, producing a depth map. RGB-D data is used for scene reconstruction and 3D modeling, object recognition and scene segmentation, and augmented reality and virtual reality. Some works refer to RGB-D data as 2.5D data, which indicates that it lies between traditional 2D images and complete 3D data. While RGB-D data may not represent a complete three-dimensional dataset in the conventional sense, its inclusion of depth information gives it some three-dimensional characteristics; here, we justify its classification as 3D data. RGB-D data can be transformed into point clouds, further integrating it into the 3D data framework.

\vspace{2mm}\noindent\textbf{Point Clouds}: are collections of points in 3D space, where each point represents a specific position and may include additional attributes such as color and intensity. Point clouds are commonly used to describe the surfaces of objects or environments in a scene. Point cloud data can be obtained through various methods, including 3D scanning with LiDAR sensors, structured light scanners, stereo vision systems, or photogrammetry techniques. Point clouds have many applications, including object identification and classification, environment mapping, localization, and obstacle detection in robotics and autonomous vehicles. Additionally, point clouds can be transformed into other 3D data formats such as voxels and meshes, enhancing their versatility in various applications.

\vspace{2mm}\noindent\textbf{Voxel}: data represents three-dimensional objects or scenes using a regular grid of volumetric pixels called voxels. Each voxel represents a small volume element in three-dimensional space and can store various attributes such as color, density, or material properties. Voxel data is typically employed in medical imaging, computer graphics, physics simulations, and computational modeling.

\vspace{2mm}\noindent\textbf{Meshes}: represent 3D objects or surfaces using a collection of vertices, edges, and faces, forming a network of interconnected triangles or polygons. Each vertex defines a point in 3D space, and each face consists of a set of vertices that define a surface. Computer graphics, animation, and simulation applications utilize Meshes. It can be obtained through 3D scanning, computer-aided design (CAD) modeling, or procedural generation. Mesh data is generally used in object segmentation, computer graphics and animation, and finite element analysis domains.

\vspace{-2mm}


\begin{figure*}[t]
\centering
\includegraphics[width=0.9\textwidth]{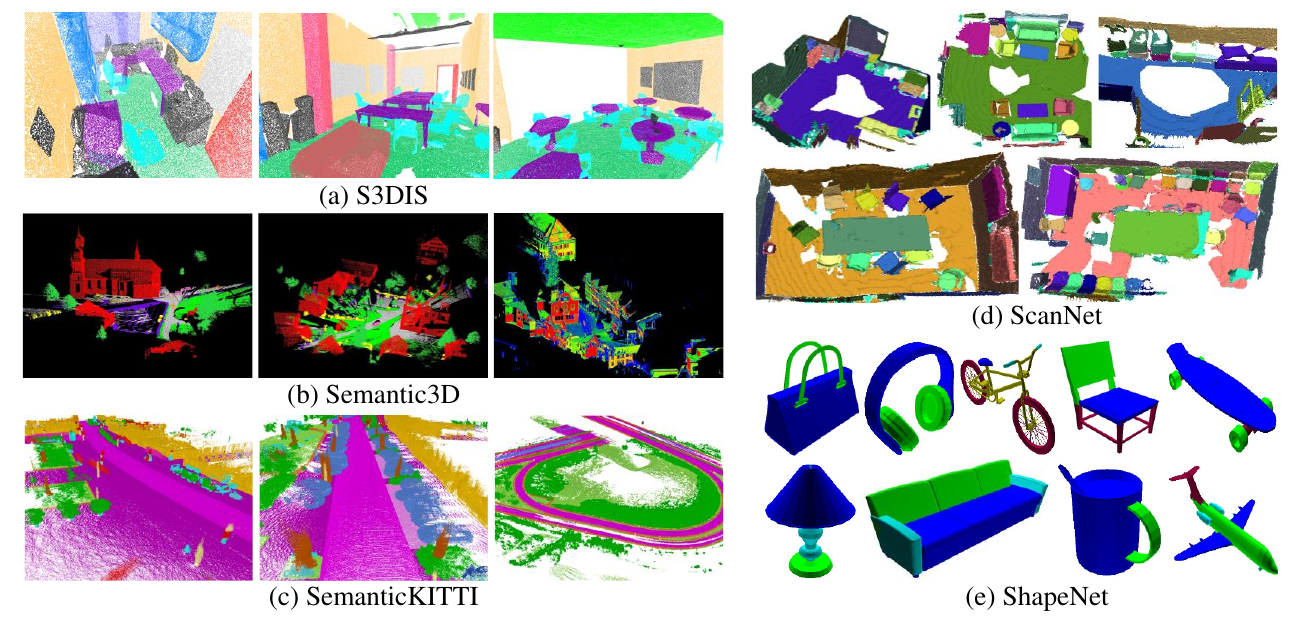}
\caption{Annotated examples from (a) S3DIS, (b) Semantic3D, (c) SemanticKITTI for 3D semantic segmentation, (d) ScanNet for 3D instance segmentation, and (e) ShapeNet for 3D part segmentation. See Table \ref{table1} for a summary of these datasets.}
\label{dataset}
\end{figure*}

\subsection{Datasets}
Datasets are critical to train and test 3D segmentation algorithms using deep learning. However, privately gathering and annotating datasets is cumbersome and expensive as it needs domain expertise, high-quality sensors and processing equipment. Thus, building on public datasets is an ideal way to reduce costs. Following this way has another advantage for the community: it compares algorithms fairly. Table~\ref{table1} summarizes some of the most popular and typical datasets concerning the sensor type, data size and format, scene class and annotation method.

These datasets are acquired for 3D semantic segmentation by different type of sensors, including RGB-D cameras~\cite{silberman2011indoor},~\cite{silberman2012indoor},~\cite{song2015sun}, \cite{hua2016scenenn}, \cite{dai2017scannet}, mobile laser scanner~\cite{roynard2018paris},~\cite{behley2019semantickitti}, static terrestrial scanner~\cite{hackel2017semantic3d} and unreal engine~\cite{brodeur2017home}, \cite{wu2018building} and other 3D scanners~\cite{armeni20163d},~\cite{chang2017matterport3d}. Among these, the ones obtained from unreal engines are synthetic datasets~\cite{brodeur2017home}, \cite{wu2018building} that do not require expensive equipment or annotation time. These are also rich in categories and quantities of objects. Synthetic datasets have complete 360-degree 3D objects with no occlusion effects or noise compared to the real-world datasets, which are noisy and contain occlusions \cite{silberman2011indoor}, \cite{silberman2012indoor}, \cite{song2015sun}, \cite{hua2016scenenn}, \cite{dai2017scannet}, \cite{roynard2018paris},~\cite{behley2019semantickitti}, \cite{armeni20163d}, \cite{hackel2017semantic3d}, \cite{chang2017matterport3d}. For 3D instance segmentation, there are limited 3D datasets, such as ScanNet~\cite{dai2017scannet} and S3DIS~\cite{armeni20163d}. These two datasets contain separate scans of real-world indoor scenes obtained by RGB-D cameras or  Matterport. For 3D part segmentation, the Princeton Segmentation Benchmark (PSB)~\cite{chen2009benchmark}, COSEG~\cite{wang2012active} and ShapeNet~\cite{yi2016scalable} are three of the most popular datasets. Below, we introduce five famous segmentation datasets in detail, including S3DIS~\cite{armeni20163d}, ScanNet~\cite{dai2017scannet}, Semantic3D~\cite{hackel2017semantic3d}, SemanticKITTI~\cite{chang2017matterport3d} and ShapeNet~\cite{yi2016scalable}. Some examples with annotation from these datasets are shown in Figure~\ref{dataset}.

\vspace{1mm}\noindent\textbf{S3DIS:} In this dataset, the complete point clouds are obtained without any manual intervention using the Matterport scanner. The dataset consists of 271 rooms belonging to six large-scale indoor scenes from three different buildings (a total of 6020 square meters). These areas mainly include offices, educational and exhibition spaces, conference rooms etc.

\vspace{1mm}\noindent\textbf{Semantic3D} comprises around four billion 3D points acquired with static terrestrial laser scanners, covering up to 160$\times$240$\times$30 meters in real-world 3D space. Point clouds belong to 8 classes (e.g., urban and rural) and contain 3D coordinates, RGB information, and intensity. Unlike 2D annotation strategies, 3D data labeling is readily amenable to over-segmentation, where each point is individually assigned to a class label.

\vspace{1mm}\noindent\textbf{SemanticKITTI} is a large outdoor dataset containing detailed point-wise annotation of 28 classes. Building on the KITTI vision benchmark~\cite{geiger2012we}, SemanticKITTI contains annotations of all 22 sequences of this benchmark consisting of 43K scans. Moreover, the dataset contains labels for the complete horizontal 360 field-of-view of the rotating laser sensor.

\vspace{1mm}\noindent\textbf{ScanNet} dataset is particularly valuable for research in scene understanding as its annotations contain estimated calibration parameters, camera poses, 3D surface reconstruction, textured meshes, dense object-level semantic segmentation, and CAD models. The dataset comprises annotated RGB-D scans of real-world environments. There are 2.5M RGB-D images in 1513 scans acquired in 707 distinct places. After RGB-D image processing, human intelligence tasks were annotated using the Amazon Mechanical Turk.

\vspace{1mm}\noindent\textbf{ShapeNet} dataset has a novel scalable method for efficient and accurate geometric annotation of massive 3D shape collections. The novel technical innovations explicitly model and lessen the human cost of the annotation effort. Researchers create detailed point-wise labeling of 31963 models in shape categories in ShapeNetCore and combine feature-based classifiers, point-to-point correspondences, and shape-to-shape similarities into a single CRF optimization over the network of shapes.

\begin{table*}[htbp]
\caption{Summary of popular 3D segmentation datasets, including the sensor, type, size, object class, number of classes (shown in brackets), and annotation method. ``S" means synthetic environment, while ``R" means real-world environment. Similarly, the symbols ``Kf", ``s," and ``Mp" stands for thousand frames, can and million points. The symbol '--' means information is unavailable.}
\label{table1}
\centering
\renewcommand\arraystretch{1.2}
\setlength{\tabcolsep}{0.5mm}{\begin{tabular*}{\textwidth}{l|l|c|c|l|l}

\specialrule{1pt}{0pt}{0pt}
\textbf{Datasets}   &\textbf{Sensors}           &\textbf{Type} &\textbf{Size}    &\textbf{Scene class (number)}               &\textbf{Annotation method}                    \\
\hline
\multicolumn{6}{c}{Datasets for 3D semantic segmentation}  \\
\hline
NYUv1~\cite{silberman2011indoor}           & Microsoft Kinect v1               & R    & 2347f   & bedroom, cafe,   kitchen, etc. (7)        & Condition Random Field model  \\
NYUv2~\cite{silberman2012indoor}          & Microsoft Kinect v1               & R    & 1449f   & bedroom, cafe, kitchen,   etc. (26)       & 2D annotation from AMK  \\
SUN RGB-D~\cite{song2015sun}     &Xtion LIVE PRO, MKv1/2 & R    & 10355f  & objects, room   layouts, etc.(47)         & 2D/3D polygons+3D Bbox      \\
SceneNN~\cite{hua2016scenenn}      & Asus Xtion PRO, MK v2             & R    & 100s    & bedroom, office,   apartment, etc.(-)     & 3D Labels project to 2D frames        \\
RueMonge2014~\cite{riemenschneider2014learning}   & --                                & R    & 428s    & window, wall,   balcony, door, etc(7)     & Multi-view semantic label.+CRF \\
ScanNet~\cite{dai2017scannet}       & Occipital structure sensor        & R    & 2.5Mf   & office,   apartment, bathroom, etc(19)    & 3D labels project to 2D frames        \\
S3DIS~\cite{armeni20163d}        & Matterport camera                & R    & 70496f  & conference rooms,   offices, etc(11)      & Hierarchical labeling                 \\
Semantic3D~\cite{hackel2017semantic3d}    & Terrestrial laser scanner         & R    & 1660Mp  & farms, town hall, sports fields, etc (8) & Three baseline methods                \\
NPM3D~\cite{roynard2018paris}          & Velodyne HDL-32E LiDAR            & R    & 143.1Mp & ground, vehicle,   hunman, etc (50)       & Human labeling                        \\
SemanticKITTI~\cite{behley2019semantickitti} & Velodyne HDL-64E                  & R    & 43Ks    & ground, vehicle,   hunman, etc(28)        & Multi-scans semantic labelling      \\
Matterport3D~\cite{chang2017matterport3d}  & Matterport camera                 & R    & 194.4Kf & various rooms   (90)                      & Hierarchical   labeling               \\
HoME~\cite{brodeur2017home}          & Planner5D platform                & S    & 45622f  & rooms, object and   etc.(84)              & SSCNet+short text description    \\
House3D~\cite{wu2018building}       & Planner5D platform                & S    & 45622f  & rooms, object and   etc.(84)              & SSCNet+3 ways                         \\\hline
\multicolumn{6}{c}{Datasets for 3D instance segmentation}                                                                                                                  \\\hline
ScanNet~\cite{dai2017scannet}       & Occipital structure sensor        & R    & 2.5Mf   & office,   apartment, bathroom, etc(19)    & 3D labels project to 2D frames        \\
S3DIS~\cite{armeni20163d}         &Matterport camera                 & R    & 70496f  & conference rooms,   offices, etc(11)      & Active learning   method              \\\hline
\multicolumn{6}{c}{Datasets for 3D part segmentation}                                                                                                                      \\\hline
ShapeNet~\cite{yi2016scalable}      & --                                & S    & 31963s  & transportation,   tool, etc.(16)          & Propagat. human   label to shapes   \\
PSB~\cite{chen2009benchmark}          & Amazon’s Mechanical Turk          & S    & 380s     & human,cup,   glasses airplane,etc(19)     & Interactive segmentation   tool       \\
COSEG~\cite{wang2012active}          & --                                & S    & 1090s    & vase, lamp,   guiter, etc (11)            & semi-supervised   learning    \\  
\specialrule{1pt}{0pt}{0pt}
\end{tabular*}}
\end{table*}



\subsection{Evaluation Metrics}

Different evaluation metrics can assert the validity and superiority of segmentation methods, including the execution time, memory footprint and accuracy. However, few authors provide detailed information about the execution time and memory footprint of their methods. This paper introduces the accuracy metrics mainly.

For 3D semantic segmentation, Overall Accuracy (OA), mean class Accuracy (mA) and mean class Intersection over Union (mIoU) are the most frequently used metrics to measure the accuracy of segmentation methods. For the sake of explanation, we assume that there are a total of $K$ classes, and $  p_{ij}$ is the {minimum unit (e.g., pixel, voxel, mesh, point)} of class $  i$ implied to belong to class $j$. In other words, $   p_{ii}$ represents true positives, while $  p_{ij}$ and $  p_{ji}$ represent false positives and false negatives, respectively.  

\noindent\textbf{Overall Accuracy}: is a straightforward metric that computes the ratio between the number of truly classified samples and the total number of samples. 
\begin{equation}
OA={\frac{\sum_{i=0}^Kp_{ii}}{\sum_{i=0}^K\sum_{j=0}^Kp_{ij}}}.
\end{equation}

\noindent\textbf{Mean Accuracy}: is an extension of OA, computing OA in a per-class and then averaging over the total number of classes.
\vspace{-2mm}
\begin{equation}
mA=\frac{1}{K+1}\sum_{i=0}^K\frac{p_{ii}}{\sum_{j=0}^Kp_{ij}}
\end{equation}

\noindent\textbf{Mean Intersection over Union}: is a standard metric for semantic segmentation. It computes the 
intersection ratio between ground truth and predicted value averaged over the total number of classes $K$.
\begin{equation}
mIoU=\frac{1}{K+1}\sum_{i=0}^K\frac{p_{ii}}{\sum_{j=0}^Kp_{ij}+\sum_{j=0}^Kp_{ji}-p_{ii}}
\end{equation}

For 3D instance segmentation, Average Precision (AP) and mean class Average Precision (mAP) are also frequently used. Assuming $  L_{I}, I\in[0, K]$ instance in every class, and $  c_{ij}$ is the amount of point of instance $  i$ inferred to belong to instance $  j$ ($  i=j$ represents correct and $  i\ne j$ represents incorrect segmentations).

\noindent\textbf{Average Precision:} is another simple metric for segmentation that computes the ratio between true positives and the total number of positive samples.
\begin{equation}
AP=\sum_{I=0}^K\sum_{i=0}^{L_{I}}\frac{c_{ii}}{c_{ii}+{\sum_{j=0}^{L_{I}}c_{ij}}}
\end{equation}
\noindent\textbf{Mean Average precision}: is an extension of AP which computes per-class AP and then averages over the total number of classes $K$.
\begin{equation}
mAP=\frac{1}{K+1}\sum_{I=0}^K\sum_{i=0}^{L_{I}}\frac{c_{ii}}{c_{ii}+{\sum_{j=0}^{L_{I}}c_{ij}}}
\end{equation}

For 3D part segmentation, the overall average category Intersection over Union ($ \rm mIoU_{cat}$) and overall average instance Intersection over Union ($ \rm  mIoU_{ins}$) are most frequently used. For the sake of explanation, we assume $   M_{J}, J \in[0, L_I]$  parts in every instance, and $  p_{ij}$ as the total number of points in part $  i$ inferred to belong to part $j$. 

\noindent\textbf{Overall average category Intersection over Union}: is an evaluation metric for part segmentation that measures the mean IoU averaged across K classes.
\begin{equation}
\small{mIoU_{cat}=\frac{1}{K+1}\sum_{I=0}^K\sum_{J=0}^{L_{I}}\sum_{i=0}^{M_{J}}\frac{p_{ii}}{\sum_{j=0}^{M_{j}}p_{ij}+\sum_{i=0}^{M_{j}}p_{ji}-p_{ii}} }
\end{equation}   

\noindent\textbf{Overall average instance Intersection over Union:} for part segmentation measures the mean IoU across all instances.
\begin{equation}
\small\small mIoU_{ins}=\frac{1}{\sum_{I=0}^KL_{I}+1}\sum_{I=0}^K\sum_{J=0}^{L_{I}}\sum_{i=0}^{M_{J}}\frac{p_{ii}}{\sum_{j=0}^{M_{j}}p_{ij}+\sum_{i=0}^{M_{j}}p_{ji}-p_{ii}}
\end{equation}


\section{3D Semantic Segmentation}\label{section3}
\begin{figure*}[t]
\centering
\includegraphics[width=0.95 \textwidth]{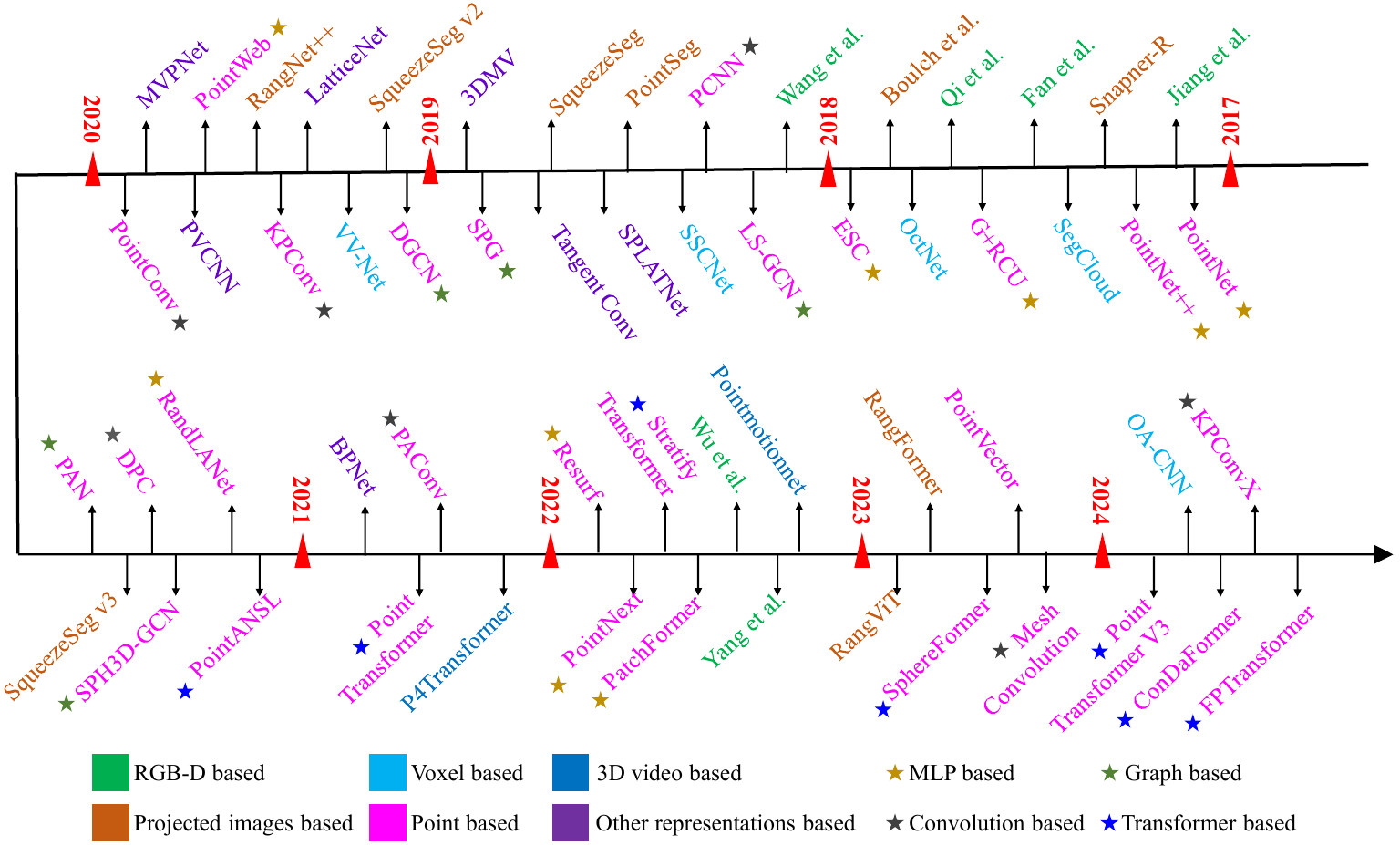}
\caption{ Milestones of deep learning based 3D semantic segmentation methods. Note that the arrow (timeline) goes anti-clockwise}
\label{milestones}
\end{figure*}

Many deep learning methods for 3D semantic segmentation have been proposed in the literature. These methods can be divided into five categories according to the data representation used, namely, RGB-D image-based, projected images-based, voxel-based, point-based, 3D video, and other representations-based. Based on the network architecture, point-based methods can be further categorized into multiple-layer perceptron (MLP) based, point convolution based, graph convolution based and point transformer based methods. Figure~\ref{milestones} shows the milestones of deep learning on 3D semantic segmentation in recent years.

\subsection{RGB-D Based Segmentation}

\begin{figure*}[t]
\centering
\includegraphics[width=\textwidth]{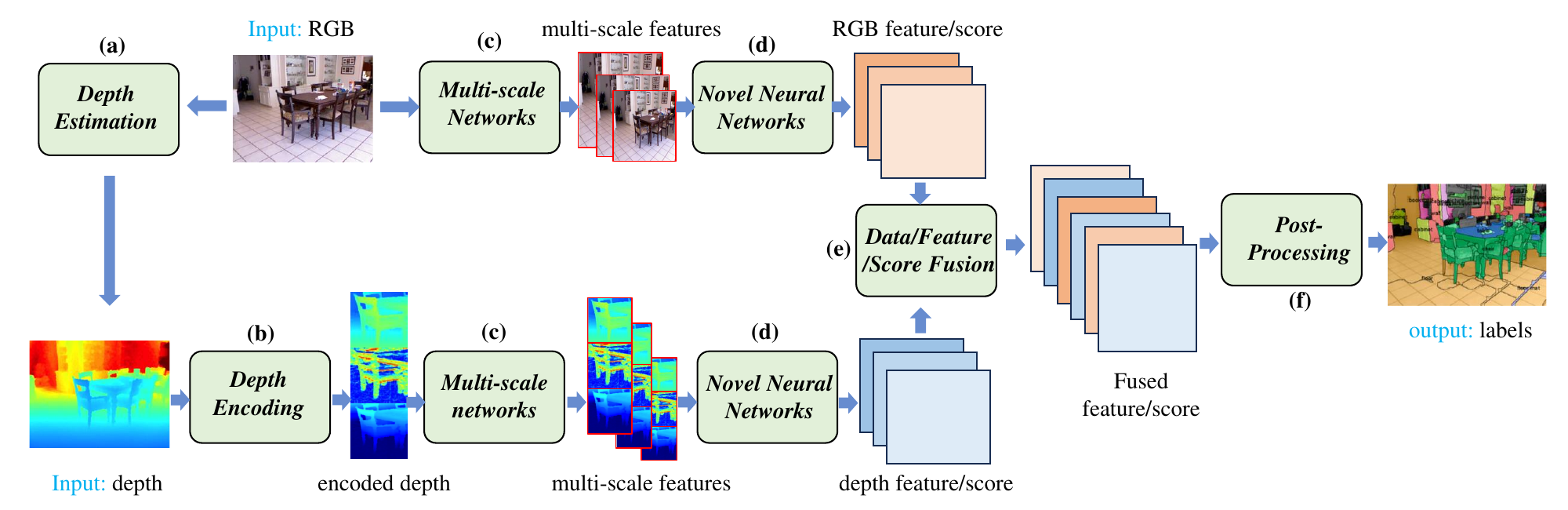}
\caption{Typical two-channel framework with six improvement modules, including (a) multi-task learning, (b) depth encoding, (c) multi-scale network, (d) novel neural network architecture, (e)~feature/score level fusion, and (f) post-processing}
\label{RGB_overview}
\end{figure*}

The depth map in an RGB-D image contains geometric information about the real world, which helps distinguish foreground objects from the background, providing opportunities to improve segmentation accuracy. In this category, the classical two-channel network is generally \textcolor{black}{employed to separately extract features from both RGB and depth images.} However, \textcolor{black}{this straightforward framework often lacks the capacity to capture detailed and comprehensive features. To address this limitation, researchers have incorporated various additional modules into this basic two-channel framework to enhance its performance} by learning more complex contextual and geometric information \textcolor{black}{for segmentation accuracy}. These modules can \textcolor{black}{generally be categorized into six groups}: multi-task learning, depth encoding, multi-scale networks, novel neural network architectures, data/feature/score level fusion and post-processing techniques (see Figure~\ref{RGB_overview}). Table ~\ref{table2} summarizes the semantic segmentation methods based on RGB-D images.

\vspace{2mm}\noindent\textbf{Multi-tasks learning:} Depth estimation and semantic segmentation \textcolor{black}{are both complex and challenging tasks} in computer vision. \textcolor{black}{These tasks are interrelated because the depth changes within a single object are generally smaller than the depth changes between different objects. Therefore,} many researchers combine depth estimation with semantic segmentation. \textcolor{black}{Based on the relationship between these two tasks, there are primarily two types of} multi-task learning frameworks: cascade and parallel.


\textcolor{black}{In the cascade framework, the depth estimation task first produces depth images, which are then used by the semantic segmentation task.} For example, Cao et al.~\cite{cao2016exploiting} \textcolor{black}{applied deep convolutional neural fields, as introduced in~\cite{liu2015learning}, to estimate depth.} The resulting depth and RGB images are then input into a two-channel FCN for semantic segmentation. Similarly, Guo et al.~\cite{guo2018semantic} \textcolor{black}{used the deep network developed by Ivanecky~\cite{ivanecky2016depth} to automatically generate depth images from a single RGB image and subsequently proposed a two-channel FCN model that utilizes the} RGB image and the predicted depth map for pixel-level labeling.


The cascade framework  \textcolor{black}{handles depth estimation and semantic segmentation separately,  preventing simultaneous end-to-end training for both tasks. As a result, the semantic segmentation task does not contribute to improving the depth estimation task. On the other hand, the parallel framework integrates both tasks within a unified network, allowing them to mutually enhance each other.} For example, Wang et al.~\cite{wang2015towards}  \textcolor{black}{employed a Joint Global CNN to leverage pixel-level depth values and semantic labels from RGB images, providing accurate global scale and semantic guidance.} They also used a Joint Region CNN to extract region-level depth values and semantic maps from RGB images,  \textcolor{black}{enabling the learning of fine-grained depth and semantic boundaries. A multi-scale FCN~\cite{mousavian2016joint} consists of five streams that capture depth and semantic features at various scales}, where both tasks share the underlying feature representations. Liu et al.~\cite{liu2018collaborative} proposed a collaborative deconvolutional neural network to jointly model these two tasks. However, the depth maps estimated from RGB images tend to be of lower quality compared to those obtained directly from depth sensors. As a result, this multi-task learning approach has gradually fallen out of favor in RGB-D semantic segmentation.


\vspace{2mm}\noindent\textbf{Depth Encoding:} \textcolor{black}{Traditional 2D CNNs cannot effectively capture} the rich geometric features from raw depth images. An alternative approach is to transform raw depth images into representations better suited for 2D CNNs. Hoft et al.~\cite{hoft2014fast} used a simplified version of the histogram of oriented gradients to represent the depth channels from RGB-D scenes. Gupta et al.~\cite{guptalearning} and Aman et al.~\cite{lin2017cascaded} \textcolor{black}{derived three new channels from the raw depth images, including horizontal disparity, height above ground, and angle with gravity (HHA).} Liu et al.~\cite{liu2018rgb} \textcolor{black}{identified a limitation of HHA, noting that some scenes may lack sufficient horizontal and vertical planes. To address this problem}, they propose a novel gravity direction detection method using vertical lines to improve representation learning. Hazirbas et al.~\cite{hazirbas2016fusenet} also argued that the HHA representation has a high computational cost and contains less information than raw depth images. \textcolor{black}{They introduced an architecture called FuseNet, which has two encoder-decoder branches—a depth branch and an RGB branch—that directly encode depth information while reducing computational complexity.}


\vspace{2mm}\noindent \textbf{Multi-scale Network:} The contextual information learned by multi-scale networks \textcolor{black}{is particularly beneficial for segmenting small objects and detailed regions}. To directly extract features from both RGB and depth images, Couprie et al.~\cite{couprie2013indoor} \textcolor{black}{used a multi-scale convolutional network.} Similarly, Aman et al.~\cite{raj2015multi} \textcolor{black}{developed a multi-scale deep ConvNet} for segmentation, where the coarse predictions from the VGG16-FC network are upsampled in a Scale-2 module and concatenated \textcolor{black}{with the low-level predictions from the VGG-M network in a Scale-1 module, capturing both high-level and low-level features. }However, this approach is sensitive to clutter in the scene, leading to output errors. Lin et al. ~[\textcolor{cyan}{107}] addressed this by focusing on low-resolution regions with higher depth and high-resolution regions with lower depth. They use depth maps to divide the corresponding color images into multiple scene-resolution regions and introduce a \textcolor{black}{context-aware receptive field (CaRF) to concentrate on the semantic segmentation of specific scene-resolution regions, making their approach a multi-scale network.}

\vspace{2mm}\noindent\textbf{Novel Neural Networks:}~Given the fixed grid computation of CNNs, their ability to process and exploit geometric information is limited. Therefore, researchers have proposed other novel neural network architectures to exploit better geometric features and the relationships between RGB and depth images. These architectures can be divided into five main categories.

\begin{itemize}
    \item 


\emph{Improved 2D Convolutional Neural Networks (2D CNNs)}. Inspired by cascaded feature networks~\cite{lin2017cascaded}, the Dense-Sensitive Fully Convolutional Neural Network (DFCN)~\cite{jiang2017incorporating} integrates depth information into the early layers of the network using \textcolor{black}{feature fusion techniques, followed by} several dilated convolutional layers to capture contextual information. Similarly, the depth-aware 2D CNN~\cite{wang2018depth} \textcolor{black}{introduced a depth-aware convolution layer and a depth-aware pooling layer, designed with the concept that pixels sharing the same semantic label and similar depth should exert a stronger influence on each other.}

\item 

\emph{Deconvolutional Neural Networks} (DeconvNets). \textcolor{black}{They provide a }simple, yet effective and efficient solution for refining the segmentation map. Liu et al.~\cite{liu2018collaborative} and Wang et al.~\cite{wang2016learning} \textcolor{black}{use the DeconvNet due to its strong performance.} However, the potential of DeconvNets is limited as the high-level prediction map mainly gathers large-scale contexts for dense prediction. \textcolor{black}{To address this problem,} the Locality-Sensitive DeconvNet (LS-DeconvNet) \cite{cheng2017locality} refines boundary segmentation over both depth and color images. LS-DeconvNet \textcolor{black}{integrates} local visual and geometric cues from the raw RGB-D data into each DeconvNet, \textcolor{black}{allowing it to upsample coarse convolutional maps with extensive contexts while preserving accuracy object boundaries.}

\item 

\emph{Recurrent Neural Networks} (RNNs) can capture long-range dependencies between pixels but are primarily designed for a single data channel, such as RGB. \textcolor{black}{Fan et al.~\cite{fan2017rgb} extended traditional RNNs to support multiple modalities, creating multimodal RNNs (MM-RNNs) for applications like RGB-D scene labeling. MM-RNNs enable the 'memory' to be shared across both color and depth channels, allowing each channel to incorporate features and attributes from the others, thereby enhancing the discriminative power of the learned features for semantic segmentation. Additionally,} a novel Long Short-Term Memorized Context Fusion (LSTM-CF) model~\cite{li2016lstm} is introduced to effectively capture and integrate contextual information from multiple RGB and depth image channels.

\item 

\emph{Graph Neural Networks}  (GNNs) are first applied to RGB-D semantic segmentation by Qi et al.~\cite{qi20173d}, who projected the 2D RGB pixels into a 3D space using depth information and linked the 3D points with their corresponding semantic data. They \textcolor{black}{then constructed a graph using the k-nearest neighbors of these 3D points} and employed a 3D Graph Neural Network (3DGNN) to \textcolor{black}{make predictions for each pixel.}

  \item 
\emph{Transformers} gained popularity in RGB image segmentation and have been extended to RGB-D segmentation. Researchers have proposed various approaches to leveraging transformers for this purpose. One notable work~\cite{ying2022uctnet} introduces the concept of uncertainty-aware self-attention, which explicitly manages the information flow from unreliable depth pixels to confident depth pixels during feature extraction. This approach addresses the challenges posed by noisy or uncertain depth information in RGB-D segmentation. Another study~\cite{wu2022transformer} adopts the Swin-Transformer directly to exploit RGB and depth features. By leveraging the self-attention mechanism, this approach captures long-range dependencies and enables effective fusion of RGB and depth information for segmentation. The success of the Swin-Transformer inspires a hierarchical Swin-RGBD Transformer~\cite{yang2022hierarchical}, which incorporates and leverages depth information to complement and enhance the ambiguous and obscured features in RGB images. The hierarchical architecture allows for multi-scale feature learning and enables more effective RGB and depth information integration.
\end{itemize}

\begin{table*}[t]
\caption{The summary of RGB-D based methods with deep learning. Est.$\leftarrow$depth estimation. Enc.$\leftarrow$depth encoding. Mul.$\leftarrow$multi-scale networks. Nov.$\leftarrow$novel neural networks. Fus.$\leftarrow$data/feature/score fusion. Pos.$\leftarrow$post-processing. 2S$\leftarrow$ 2-stream.}
\vspace{2mm}
\label{table2}
\scriptsize
\centering
\renewcommand\arraystretch{1.4}
\setlength{\tabcolsep}{2mm}
{\begin{tabular*}{0.98\textwidth}{l|llllll|l|l}
\specialrule{1pt}{0pt}{0pt}
\textbf{Methods}    &Est. &Enc. &Mul. &Nov. &Fus. &Pos. & \textbf{Archi. (2S)}   & \textbf{Contribution}                                                                                 \\  \hline
Cao et al~\cite{cao2016exploiting}     &$\checkmark$    &$\checkmark$    &$\times$    &$\times$    &$\checkmark$    &$\times$    & FCNs                  & Estimating depth images+a unified network for two tasks                                                            \\
Guo et al.~\cite{guo2018semantic}    &$\checkmark$    &$\times$    &$\times$    &$\times$    &$\checkmark$    &$\times$    & FCNs                  & Incorporating depth \& gradient for depth estim.   \\
Wang et al.~\cite{wang2015towards}     &$\checkmark$    &$\times$    &$\times$    &$\times$    &$\times$    &$\checkmark$    & Region./Global   CNN  & HCRF for   fusion and refining + two tasks by a network                                      \\
Mousavian et al.~\cite{mousavian2016joint}  &$\checkmark$    &$\times$    &$\checkmark$    &$\times$    &$\checkmark$    & $\checkmark$    & FCN                   & FC-CRF for   refining + Mutual improvement for two tasks                                     \\
Liu et al.~\cite{liu2018collaborative}      &$\checkmark$    &$\times$    &$\times$    &$\checkmark$    &$\times$    &$\checkmark$    & S/D-DCNN               & PBL for two   feature maps integration + FC-CRF                         \\
Hoft et al.~\cite{hoft2014fast}    &$\times$    &$\checkmark$    &$\times$    &$\times$    &$\times$    &$\times$    & CNNs                  & A embedding   for depth images                                                               \\
Gupta et al.~\cite{guptalearning}    &$\times$    &$\checkmark$    &$\times$    &$\times$    &$\times$    &$\times$    & CNNs                  & HHA for depth images                                                                         \\
Liu et al.~\cite{liu2018rgb}     &$\times$    &$\checkmark$    &$\times$    &$\times$    &$\checkmark$    &$\checkmark$    & DCNNs                 & New depth   encoding+ FC-CRF for refining                                                  \\
Hazirbas et al.~\cite{hazirbas2016fusenet} &$\times$    &$\checkmark$    &$\times$    &$\times$    &$\checkmark$    &$\times$    & Encoder-decoder       & Semantic   and depth feature fusion at each layer                                            \\
Coupri et al.~\cite{couprie2013indoor} &$\times$    &$\times$    &$\checkmark$    &$\times$    &$\checkmark$    &$\times$    & ConvNets              & RGB laplacian pyramid for multi-scale   features                                             \\
Raj et al.~\cite{raj2015multi}     &$\times$    &$\checkmark$    &$\checkmark$    &$\times$    &$\checkmark$    &$\times$    & VGG-M                 & New multi-scale deep CNN                                                                   \\
Lin et al.~\cite{lin2017cascaded}      &$\times$    &$\times$    &$\checkmark$    &$\checkmark$    &$\checkmark$    &$\times$    & CFN                   & CaRF for multi-resolution features                                                           \\
Jiang et al.~\cite{jiang2017incorporating}  &$\times$    &$\times$    &$\times$    &$\checkmark$    &$\checkmark$    &$\checkmark$    & RGB-FCN               & Semantic \& depth feature fusion at each layer + DCRF                                         \\
Wang et al.~\cite{wang2018depth}     &$\times$    &$\times$    &$\times$    &$\checkmark$    &$\times$    &$\times$    & Depth-aware   CNN     & Depth-aware   Conv. and depth aware average pooling                                          \\
Cheng et al.~\cite{cheng2017locality}  &$\times$    &$\checkmark$    &$\times$    &$\checkmark$    &$\checkmark$    &$\times$    & FCN + Deconv          & LS-DeconvNet + novel gated fusion                                                           \\
Fan et al.~\cite{fan2017rgb}     &$\times$    &$\times$    &$\times$    &$\checkmark$    &$\checkmark$    &$\times$    & MM-RNNs               & Multimodal   RNN                                                                             \\
Li et al.~\cite{li2016lstm}      &$\times$    &$\checkmark$    &$\times$    &$\checkmark$    &$\checkmark$    &$\times$    & LSTM-CF               & LSTM-CF for capturing and fusing   contextual inf.                                           \\
Qi et al.~\cite{qi20173d}       &$\times$    &$\times$    &$\times$    &$\checkmark$    &$\times$    &$\times$    & 3DGNN                 & GNN for RGB-D semantic segmentation                                                          \\
Wang et al.~\cite{wang2016learning}    &$\times$    &$\times$    &$\times$    &$\checkmark$    &$\checkmark$    &$\times$    & ConvNet-DeconvNet & MK-MMD for assessing the similarity between common features  \\
Ying et al.~\cite{ying2022uctnet} &$\times$    &$\times$    &$\times$    &$\checkmark$    &$\checkmark$    &$\times$ &Swin-Transformer & Effective and scalable fusion module based on aross-attention\\
Wu et al.~\cite{wu2022transformer} &$\times$    &$\times$    &$\times$    &$\checkmark$    &$\checkmark$    &$\times$ &Swin-Transformers & Transformer-based fusion module\\
Yang et al.~\cite{yang2022hierarchical} &$\times$    &$\times$    &$\times$    &$\checkmark$    &$\times$    &$\times$ &SwinT+ResNet & Swin-RGB-D Transformer \\

\specialrule{1pt}{0pt}{0pt}
\end{tabular*}}
\end{table*}

\vspace{2mm}\noindent\textbf{Fusion:} \textcolor{black}{Achieving an optimal combination of texture (from the RGB channels) and geometric (from the depth channel) information is crucial for precise semantic segmentation. Three main fusion strategies exist: data-level, feature-level, and score-level fusion, corresponding to early, middle, and late fusion, respectively.} A basic data-level fusion approach \textcolor{black}{involves merging the RGB and depth images into a four-channel input for direct use} in a CNN model~\cite{couprie2013indoor}. However, this data-level fusion \textcolor{black}{approach fails to fully utilize the strong correlations} between the depth and \textcolor{black}{RGB} channels. \textcolor{black}{In contrast, feature-level fusion captures these correlations more effectively.} For instance, a memorized fusion layer~\cite{li2016lstm} adaptively merges vertical depth information with RGB contexts in a data-driven way, \textcolor{black}{while also allowing bidirectional propagation along the horizontal direction to maintain true 2D global contexts.}

Moreover, Wang et al.~\cite{wang2016learning} introduced a feature transformation network that establishes correlations between the depth and RGB channels and \textcolor{black}{connects} the convolutional and deconvolutional networks \textcolor{black}{within} a single channel. \textcolor{black}{This network can identify unique features within a single channel and common features across both channels, enabling the two branches to share features and thereby enhance the representation capability of the shared information.} The complex feature-level fusion models mentioned above are inserted at a specific corresponding layer between the RGB and depth channels, \textcolor{black}{which makes them difficult to train and limits the fusion of other corresponding layer features.} To address this, Hazirbas et al.~\cite{hazirbas2016fusenet} and Jiang et al.~\cite{jiang2017incorporating} perform fusion using an element-wise summation to combine features from multiple corresponding layers between the two channels. Wu et al.~\cite{wu2022transformer} propose a new transformer-based fusion approach, TransD-Fusion, to more effectively capture long-range contextual information.

Score level fusion is commonly performed using the simple averaging strategy. However, the contributions of the RGB and depth models for semantic segmentation are different. A score-level fusion layer~\cite{liu2018rgb} with a weighted summation uses a convolution layer to learn the weights from the two channels. Similarly, a gated fusion layer~\cite{cheng2017locality} learns the varying performance of RGB and depth channels for different class recognition in various scenes. Both techniques improved the results over the simple averaging strategy at the cost of additional learnable parameters.

\vspace{2mm}\noindent\textbf{Post-Processing:} The results of CNN or DCNN used for RGB-D semantic segmentation are generally very coarse, resulting in rough boundaries and the disappearance of small objects. A standard method to address this problem is to couple the CNN with a conditional random field (CRF). The joint inference of hierarchical CRF (HCRF)~\cite{wang2015towards} further boosts the mutual interactions between the two channels. It enforces synergy between global and local predictions, where the global layouts guide the local predictions and reduce local ambiguities, and local results provide detailed regional structures and boundaries. A fully connected CRF (FC-CRF) for post-processing is adopted by~\cite{mousavian2016joint},~\cite{liu2018collaborative}, and~\cite{liu2018rgb}, where the pixel-wise label prediction jointly considers geometric constraints, such as pixel-wise normal information, pixel position, intensity, and depth, to promote the consistency of pixel-wise labeling. Similarly, dense-sensitive CRF (DCRF)~\cite{jiang2017incorporating} integrates the depth information with FC-CRF.


\subsection{Projected Images Based Segmentation}

\textcolor{black}{The main idea behind} projected image-based semantic segmentation is to \textcolor{black}{leverage 2D CNNs to extract features from 2D projections of 3D scenes or shapes} and subsequently combine these features for label prediction. \textcolor{black}{This approach allows for capturing} richer semantic information from large-scale scenes compared to single-view images, and reduces the data size of a 3D scene \textcolor{black}{relative to} a point cloud. The projected images are typically multi-view or spherical images.

Among these, multi-view image projection is usually employed on RGB-D datasets~\cite{dai2017scannet} and statics terrestrial scanning datasets~\cite{hackel2017semantic3d}. Spherical image projection is generally utilized on self-driving mobile laser scanning datasets~\cite{behley2019semantickitti}. Table~\ref{table3} summarizes projected images-based semantic segmentation methods.

\begin{figure}[tbp]
\centering
\includegraphics[width=0.95\columnwidth]{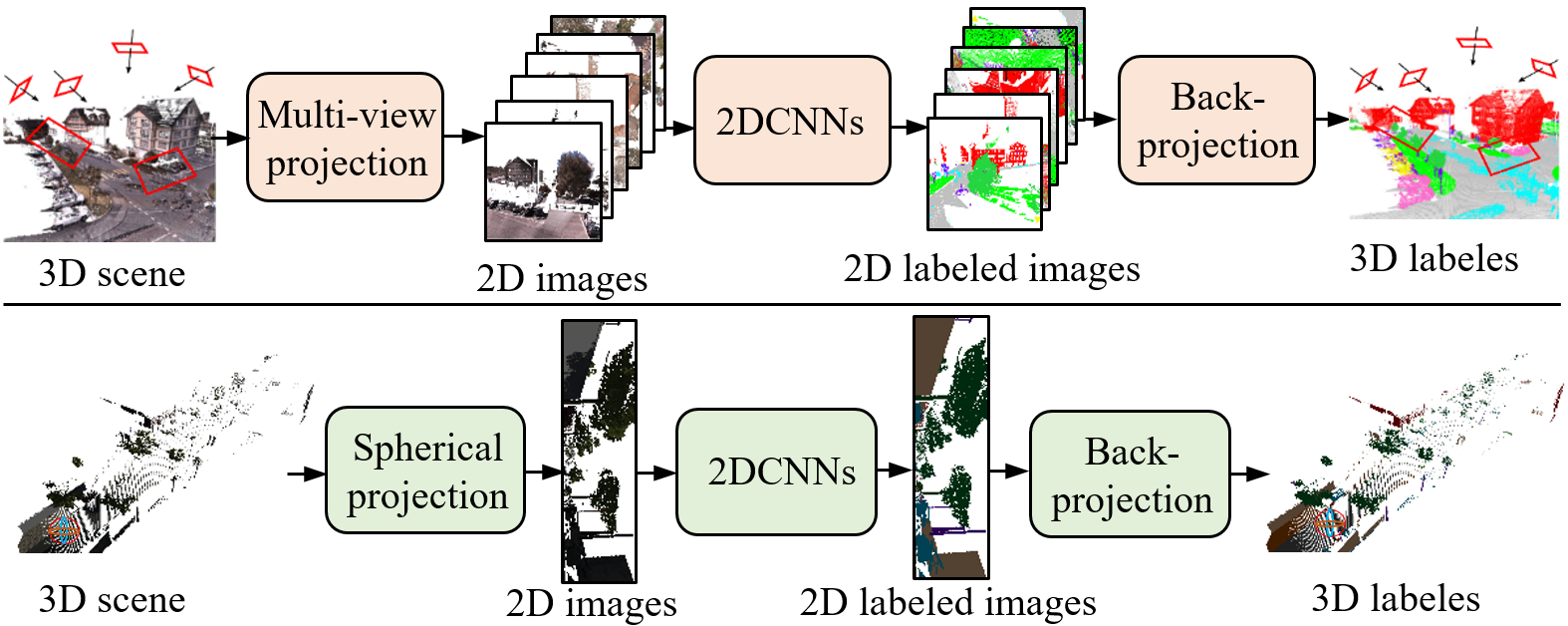}
\caption{Illustration of basic frameworks for projected images based segmentation methods. \textbf{Top}: Multi-view images based framework. \textbf{Bottom}: Spherical images based framework.}
\label{projection}
\end{figure}

\vspace{2mm}\noindent\textbf{Multi-view image segmentation:} MVCNN~\cite{su2015multi} employs a unified network to \textcolor{black}{merge features from multiple views of a 3D shape captured by a virtual camera into a single, compact shape descriptor to enhance classification performance}. This motivated researchers to apply a similar approach to 3D semantic segmentation (see Figure~\ref{projection}). For instance, Lawin et al.~\cite{lawin2017deep} project point clouds into multi-view synthetic images, such as RGB, depth, and surface normal images. The prediction scores from all multi-view images \textcolor{black}{are fused into a unified representation} and then reprojected onto each point. However, if the point cloud is sparsely populated, the snapshot may incorrectly capture points located behind the observed structure, \textcolor{black}{leading the deep network to misinterpret the various views}.

To this end, SnapNet~\cite{boulch2017unstructured},~\cite{boulch2018snapnet} preprocesses point clouds by calculating point features and creating a mesh, \textcolor{black}{similar to the process of point cloud densification.} From the mesh and point clouds, the authors generate RGB and depth images using appropriate snapshots. They then perform pixel-wise labeling of these 2D snapshots using FCNs and rapidly reproject the labels back onto the 3D points using efficient buffering. \textcolor{black}{These methods require acquiring the entire point cloud of the 3D scene beforehand to provide a complete spatial structure necessary for accurate back-projection.} However, multi-view images captured directly from real-world scenes often lose significant spatial information. Some approaches attempt to combine 3D scene reconstruction with semantic segmentation, where scene reconstruction can compensate for the missing spatial data. For instance, Guerry et al.~\cite{guerry2017snapnet} reconstruct 3D scenes using global multi-view RGB and gray scale stereo images, \textcolor{black}{then reproject the labels from 2D snapshots onto the reconstructed scene.} However, simple back-projection does not effectively integrate semantic and spatial geometric features. In response, Pham et al.~\cite{pham2019real} proposed a novel higher-order CRF, applied after back-projection, to further refine the initial segmentation.

\vspace{2mm}\noindent\textbf{Spherical image segmentation:}

Selecting snapshots from a 3D scene is challenging. \textcolor{black}{It requires careful consideration of factors such as} the number of viewpoints, the viewing distance, and the angle of the virtual cameras to optimally represent the entire scene. \textcolor{black}{To bypass} these complexities, researchers have projected the entire point cloud onto a sphere (see the last row of Figure~\ref{projection}). For example, SqueezeSeg~\cite{wu2018squeezeseg}, an end-to-end pipeline inspired by SqueezeNet~\cite{iandola2016squeezenet}, extracts features from spherical images, which are then refined by a CRF implemented as a recurrent layer. Similarly, PointSeg~\cite{wang2018pointseg} \textcolor{black}{builds on SqueezeNet by incorporating feature-wise and channel-wise attention to learn a more robust representation. SqueezeSegv2~\cite{wu2019squeezesegv2} enhances SqueezeSeg’s architecture with a context aggregation module, adding a LiDAR mask as an additional channel to improve noise robustness.} RangeNet++~\cite{milioto2019rangenet++} transfers semantic labels back to the 3D point clouds, ensuring that no points are lost regardless of the level of CNN discretization. Despite the similarities between regular RGB and LiDAR images, the feature distribution of LiDAR images varies depending on their location. SqueezeSegv3~\cite{xu2020squeezesegv3} \textcolor{black}{introduces} a spatially-adaptive and context-aware convolution, known as spatially-adaptive convolution (SAC), which applies different filters to different locations. Inspired by the success of the 2D vision Transformer, RangViT~\cite{ando2023rangevit} utilizes Vision Transformers (ViTs) pre-trained on extensive natural image datasets by adding downsampling and upsampling modules to the top and bottom of the ViTs, achieving superior performance compared to projection-based methods. Similarly, RangeFormer~\cite{kong2023rethinking} uses a scalable training strategy that divides the entire projection image into several sub-images, processes them through ViTs during training, and then sequentially merges the predictions to reconstruct the full scene.

\begin{table*}
\centering
\caption{Summary of projected images, voxel, and other-representation based methods with deep learning. M$\leftarrow$multi-view image. S$\leftarrow$spherical image. V$\leftarrow$voxel. T$\leftarrow$tangent images. L$\leftarrow$lattice. P$\leftarrow$point clouds.}
\label{table3}
\scriptsize
\renewcommand\arraystretch{1.4}
\setlength{\tabcolsep}{2mm}{\begin{tabular*}{\textwidth}{@{\extracolsep\fill}l|l|c|l|l|l}
\specialrule{1pt}{0pt}{0pt} 
&\textbf{Methods} &\textbf{Input} &\textbf{Architecture}       &\textbf{Feature extractor}       &\textbf{Contribution} \\\hline     
\multirow{12}{*}{\begin{sideways}projection\end{sideways}}
                                            & Lawin~et~al.~\cite{lawin2017deep}     & M     & multi-stream       & VGG-16                  & Investigate the impact of different   input modalities                          \\      
                                            & Boulch~et~al.~\cite{boulch2017unstructured}~\cite{boulch2018snapnet} & M     & SegNet/U-Net       & VGG-16                  & New and efficient framework SnapNet                                          \\    
                                            & Guerry~et~al.~\cite{guerry2017snapnet}    & M     & SegNet/U-Net       & VGG-16                  & Improved MVCNN+3D consistent data augment.         \\    
                                            & Pham~et~al.~\cite{pham2019real}       & M     & Two-stream         & 2DConv                  & High-order CRF+ real-time   reconstruction pipeline                             \\    
                                            & Wu~et~al.~\cite{wu2018squeezeseg}         & S     & AlexNet            & Firemodules             & End-to-end pipeline SqueezeSeg + real time                                      \\    
                                            & Wang~et~al.~\cite{wang2018pointseg}       & S     & AlexNet            & Firemodules             & Quite light-weight framework PointSeg   + real time                             \\    
                                            & Wu~et~al.~\cite{wu2019squeezesegv2}      & S     & AlexNet            & Firemodules             & Robust framework SqueezeSegV2                                                 \\    
                                            & Milioto~et~al.~\cite{milioto2019rangenet++}   & S     & DarkNet            & Residual block          & GPU-accelerated   post-processing + RangNet++                                    \\    
                                            & Xu~et~al.~\cite{xu2020squeezesegv3}         & S     & RangeNet           & SAC                     & Adopting different filters for   different locations                            \\ 
                                            & Ando~et~al.~\cite{ando2023rangevit}         & S     & U-Net           & ViTs                    &Decreasing the gaps between image and point domain.                             \\ 
                                            & Kong~et~al.~\cite{kong2023rethinking}         & S     & U-Net           & ViTs                     & Introducing a scalable training from range view strategy                        \\ \hline     
\multirow{8}{*}{\begin{sideways}voxel\end{sideways}}
                                            & Huang~et~al.~\cite{huang2016point}     & V     & 3D CNN              & 3DConv                  & Efficiently handling large data                                                 \\     
                                            & Tchapmi~et~al.~\cite{tchapmi2017segcloud}   & V     & 3D FCNN             & 3DConv                  & Combining 3D FCNN with fine-represen.                   \\    
                                            & Meng~et~al.~\cite{meng2019vv}     & V     & VAE                & RBF                     & A novel voxel-based representation +   RBF                                      \\  
                                            & Liu~et~al.~\cite{liu20173DCNN}       & V     & 3D CNN/DQN/RNN      & 3DConv                  & Integrating three vision tasks into one frame.           \\ 
                                            & Rethage et at.~\cite{rethage2018fully}  & V     & 3D FCNN             & FPConv                  & First fully-convolutional network on raw point sets             \\   
                                            & Dai~et~al.~\cite{dai2018scancomplete}       & V     & 3D FCNN             & 3DConv                  & Combing scene completion and semantic   labeling                                \\   
                                            & Riegler~et~al.~\cite{riegler2017octnet}   & V     & Octree             & 3DConv                  & Making DL with high-resolution voxels                                           \\   
                                            & Graham~et~al.~\cite{graham20183d}     & V     & FCN/U-Net          & SSConv                  & SSConv   with less computation                                                  \\ 
                                            & Peng~et~al.~\cite{peng2024oa}     & V     & U-Net          &SSConv                  & Adaptive receptive field and adaptive relation learning                                                  \\ 
                                            \hline     
\multirow{9}{*}{\begin{sideways}others\end{sideways}}
                                            & TangentConv~\cite{tatarchenko2018tangent}    & T     & U-Net              & TConv                   & Tangent   convolution + Parsing large scenes                                    \\ 
                                            & SPLATNet~\cite{su2018splatnet}         & L     & DeepLab            & BConv                   & Hierarchical and spatially-aware feature learning                               \\     
                                            & LatticeNet~\cite{rosu2019latticenet}      & L     & U-Net              & PN+3DConv         & Hybrid architecture + novel slicing operator                                          \\  
                                            & 3DMV~\cite{dai20183dmv}             & M+V   & Cascade frame.     & ENet+3DConv             & Inferring 3D semantics from both 3D and 2D input                 \\    
                                            & Hung~et~al.~\cite{chiang2019unified}     & V+M+P & Parallel frame.    & SSCNet/DeepLab/PN       & Leveraging 2D and 3D features                                                   \\   
                                            & PVCNN~\cite{liu2019point}          & V+P  & POintNet           & PVConv                  & Both memory and computation efficient                                           \\    
                                            & MVPNet~\cite{jaritz2019multi}         & M+P   & Cascade frame.     & U-Net+PointNet++              & Leveraging 2D and 3D features                                                   \\   
                                            & LaserNet++~\cite{meyer2019sensor}       & M+P   & Cascade frame.     & ResNet+LNet         & Unified network for two tasks \\
                                            
                                            & BPNet~\cite{hu2021bidirectional}       & M+P   & Cascade frame.     & 2/3DUNet         & Bidirection projection module
                                            
                                            \\ \specialrule{1pt}{0pt}{0pt}   
\end{tabular*}}
\end{table*}


\subsection {Voxel-Based Segmentation}
Similar to pixels, voxels divide the 3D space into many volumetric grids with a specific size and discrete coordinates. It contains more geometric information about the scene compared to projected images. 3D ShapeNets~\cite{wu20153d} and VoxNet~\cite{maturana2015voxnet} take volumetric occupancy grid representation as input to a 3D convolutional neural network for object recognition, which guides 3D semantic segmentation based on voxels. Voxel-based semantic segmentation methods are summarized in Table~\ref{table3}.

3D CNN is a typical architecture that processes uniform voxels for label prediction. 3D FCN~\cite{huang2016point} is for coarse voxel-level predictions but is limited by the spatial inconsistency between predictions and provides coarse labeling. SEGCloud~\cite{tchapmi2017segcloud}, a novel network, produces fine-grained predictions, upsampling the coarse voxel-wise prediction obtained from a 3D FCN to the original 3D point space resolution by trilinear interpolation.

With fixed-resolution voxels, the computational complexity grows linearly with the increase in scene scale. Large voxels can lower the computational cost of large-scale scene parsing. Liu~et~al.~\cite{liu20173DCNN} introduced a novel network called 3D CNN-DQN-RNN. Like the sliding windows in 2D semantic segmentation, 3D CNN-DQN-RNN proposes an eye window that traverses the whole data for fast localizing and segmenting class objects under the control of a 3D CNN and a deep Q-Network (DQN). The 3D CNN and residual RNN further refine features in the eye window. The pipeline learns key features of interesting regions efficiently to enhance the accuracy of large-scale scene parsing with less computational cost. Rethage~et~al.~\cite{rethage2018fully} present a novel fully convolutional point network (FCPN), sensitive to multi-scale input, to parse large-scale scenes without pro- or post-process steps. Mainly, FCPN is able to learn memory-efficient representations that scale well to larger volumes.

Similarly, Dai~et~al.~\cite{dai2018scancomplete} design a novel 3D CNN to train on-scene subvolumes but deploy on arbitrarily large scenes at test time, as it can handle large scenes with varying spatial extent. Additionally, their network adopts a coarse-to-fine tactic to predict multiple resolution scenes to handle the resolution growth in data size as the scene increases. Traditionally, the voxel representation only comprises boolean occupancy information, which loses many geometric details. Meng~et~al.~\cite{meng2019vv} develop a novel information-rich voxel representation by using a variational auto-encoder (VAE) and a radial basis function (RBF) to capture the distribution of points within each voxel. Further, they proposed a group equivariant convolution to exploit this feature.

In fixed-scale scenes, the computational complexity grows cubically as the voxel resolution increases. However, the volumetric representation is naturally sparse, resulting in unnecessary computations when applying 3D dense convolution to the sparse data. To address this problem, OctNet~\cite{riegler2017octnet} divides the space hierarchically into nonuniform voxels using a series of unbalanced octrees. The tree structure allows memory allocation and computation to focus on relevant dense voxels without sacrificing resolution. However, empty space still imposes a computational and memory burden on OctNet. In contrast, Graham~et~al.~\cite{graham20183d} proposed a novel submanifold sparse convolution (SSC) that does not perform computations in empty regions, making up for the drawback of OctNet. OA-CNN~\cite{peng2024oa} introduces adaptive receptive fields and adaptive relation into the Sparse CNN, to learn long-scale context information.


\subsection {Point-Based Segmentation}
Point clouds are scattered irregularly in 3D space, lacking any canonical order and translation invariance, which restricts the use of conventional 2D/3D convolutional neural networks. Recently, a series of point-based semantic segmentation networks have been proposed. These methods can be roughly subdivided into four categories: MLP-based, point convolution-based, graph convolution-based, and transformer-based. These methods are summarized in Table~\ref{table4}.

\begin{table*}
\centering
\vspace{1mm}
\caption{Summary of point based semantic segmentation methods with deep learning. Here, `Nb' stands for Neighbour, `part.' is for partition.}
\vspace{1mm}
\label{table4}
\scriptsize
\renewcommand\arraystretch{1.4}
\setlength{\tabcolsep}{2mm}{\begin{tabular*}{0.95 \textwidth}{@{\extracolsep\fill}c|l|c|c|c|l}

\specialrule{1pt}{0pt}{0pt} 
& \textbf{Methods} & \textbf{Nb. Search} & \textbf{Feature abstraction} & \textbf{Coarsening} & \textbf{Contribution}                                                            \\ \hline 
\multirow{9}{*}{\begin{sideways}MLP\end{sideways}}
                             & PointNet~\cite{qi2017pointnet}   & None               & MLP                 & None              & Pioneering processing points directly                                                        \\ 
                             & G+RCU~\cite{engelmann2018know}       & None               & MLP                 & None              & Two local definition+local/global pathway \\ 
                             & ESC~\cite{engelmann2017exploring}        & None               & MLP                 & None              & MC/Grid Block for local defini.+RCUs for context                    \\ 
                             & HRNN~\cite{ye20183d}       & None               & MLP                 & None              & 3P for local feature +HRNN for local context                       \\ 
                             & PointNet++~\cite{qi2017pointnet++} & Ball/KNN            & PointNet    & FPS    & Proposing hierarchical learning framework                                                    \\ 
                             & PointSIFT~\cite{jiang2018pointsift} & KNN     & PointNet   & FPS  & PointSIFT module for local shape information  \\
                             
                             & PointWeb~\cite{zhao2019pointweb}   & KNN   & PointNet  & FPS    & AFA for interactive feature exploitation                                                     \\  
                             & Repsurf~\cite{ran2022surface}   & KNN   & PointNet  & FPS    & Local triangular orient. +  local umbrella orient.                                                      \\  

                             & PointNeXt~\cite{qian2022pointnext}   & KNN   & InvResMLP  & FPS    & Next version of PointNet \\

                             & PointVector~\cite{deng2023pointvector}   & KNN   & PointNet  & FPS    & A vector oriented point set abstraction \\

                             \hline
                             
\multirow{12}{*}{\begin{sideways}Point Convolution\end{sideways}}
                             & RSNet~\cite{huang2018recurrent}      & None               & 1x1 Conv            & None              & LDM for local context exploitation                                                           \\ 
                             & DPC~\cite{engelmann2020dilated}        & DKNN               & PointConv           & None              & Dilated KNN for expanding the receptive field                                                \\ 
                             & PointWiseCNN~\cite{hua2018pointwise}       & Grid               & PWConv.             & None              & Novel point convolution                                                                      \\ 
                             & PCCN~\cite{wang2018deep}        & KD index           & PCConv.             & None              & KD-tree index for neigh. search+novel point Conv.                                            \\ 
                             & KPConv~\cite{thomas2019kpconv}      & Ball                & KPConv.             & Grid Samp.      & Novel point convolution                                                                      \\ 
                             
                             & KPConvX~\cite{deng2023pointvector}   & Ball   & KPConX  & Grid Samp.    & Novel Point Convolution \\

                             & FlexConv~\cite{groh2018flex}   & KD index           & flexConv.           & IDISS             & Novel point Conv.+flex-maxpooling no subsampling    \\ 
                             & PointCNN~\cite{li2018pointcnn}   & DKNN               & $\chi$-Conv              & FPS               & Novel point  convolution \\
                             
                             & MCC~\cite{hermosilla2018monte} & Ball                & MCConv.        & PDS      & Novel coarsening layer+point convolution                                                     \\ 
                             & PointConv~\cite{wu2019pointconv}  & KNN    & PointConv  & FPS  & Novel point convolution considering point density                                            \\ 
                             & A-CNN~\cite{komarichev2019cnn}    & DKNN         & AConv   & FPS & Novel neighborhood search+point convolution                                                  \\ 
                             & RandLA-Net~\cite{hu2020randla} & KNN     & LocSE & RPS     & LFAM with large receptive field \&  geometric details        \\ 
                             & PolarNet~\cite{zhang2020polarnet}  & None    & PointNet   & PolarGrid   & Novel local regions definition + RingConv
                             
                             \\ \hline
\multirow{10}{*}{\begin{sideways}Graph Convolution\end{sideways}}
                             & DGCNN~\cite{wang2019dynamic}        & KNN                & EdgeConv            & None              & Novel graph convolution + updating graph                                                      \\ 
                             & SPG~\cite{landrieu2018large}         & partition          & PointNet                  & None              & Superpoint graph + parsing large-scale scene                                                 \\ 
                             & DeepGCNs~\cite{li2019deepgcns}    & DKNN               & DGConv              & RPS               & Adapting residual connections between layers                                                 \\ 
                             & SPH3D-GCN~\cite{lei2020spherical}   &Ball                   & SPH3D-GConv         &FPS             & Novel graph convolution + pooling + uppooling                                                    \\ 
                             
                             & LS-GCN~\cite{wang2018local}  & KNN   & Spec.Conv.    & FPS       & Local spectral graph + Novel graph convolution                                               \\ 
                             & PAN~\cite{feng2020point}  & Multi-direct.       & LAE-Conv       & PFS    & Point-wise spatial attention+local graph Conv.                                               \\ 
                             & TGNet~\cite{li2019tgnet} & Ball     & TGConv      & PFS       & Novel graph Conv.+multi-scale features explo.                                                \\ 
                             & HDGCN~\cite{liang2019hierarchical}  & KNN                 & DGConv     & FPS       & Depthwise graph Conv. +  Pointwise Conv.                                                       \\ 
                             & 3DCon.Net~\cite{zeng20183dcontextnet} & KNN                 & PointNet  & Tree layer  & KD tree structure                                                                            \\ 
                             &$\psi$-CNN~\cite{lei2019octree}& Octree neig.        &$\psi$-Conv     & Tree layer   & Octree structure+ Novel graph convolution   \\ \hline
                             
\multirow{10}{*}{\begin{sideways}Point Transformer\end{sideways}}
                             & PGCRNet~\cite{ma2020global}     & None               & Conv1D              & None              & PointGCR to model context dependencies  \\ 
                             & AGCN~\cite{xie2020point}        & KNN                & MLP                 & None              & Point attention layer for aggregating local features    \\
                             & PointANSL~\cite{yan2020pointasnl}        & KNN                & local-nonlocal module                 & AS              & Local-nonlocal module + adaptive sampling     \\                             & Point Transformer~\cite{zhao2021point}        & KNN                & Point Transformer                 & Maxpooling              &  MLP-based relative position encoding + vec. atten.     \\
                             & Point Transformer v2~\cite{wu2022point}        & Grid part.               & Point Transformer v2                 & Gridpooling              &  Novel position encoding + Grid pooling     \\

                             & FPTransformer~\cite{he2024full}   & KNN   & Full point transformer  & SADS    & Full point encoding + shape aware downsampling \\
                             
                             & PatchFor.~\cite{zhang2022patchformer}        & Boxes part.               & Patch Transformer                 & DWConv              & First linear attention + Lightweight multi-scale    \\
                             & Fast Point Transfor.~\cite{park2022fast}        & Voxel part.                & Fast point Transformer                 & Voxel Samp.              & Lightweight local self-attention + position encoding      \\
                             & Stratified Transfor.~\cite{lai2022stratified}        & Voxel part.              & Stratified Transformer                 & PFS              & Contextual relative position encoding      \\
                            & SphereFormer~\cite{lai2023spherical}        & Voxel part.                  & Spherefor. + cubicfor.                & Maxpooling              & Novel spherical window for LIDAR points \\

                            & ConDaFormer~\cite{duan2024condaformer}   & Voxel part.   & ConDaFormer  &Maxpooling    &  Disassembled window attention module\\

                            & Point Transformer v3~\cite{wu2024point}   & Spatial proximity   &Point Transformer v3  & Gridpooling    &  Streamlined approach tailored for serialized point clouds  \\
\specialrule{1pt}{0pt}{0pt} 
\end{tabular*}}
\end{table*}

\vspace{2mm}\noindent\textbf{MLP segmentation techniques:}
techniques apply a MLP directly to the points to learn features. The PointNet~\cite{qi2017pointnet} is a pioneering work that directly processes point clouds. It uses shared MLP to exploit pointwise features and adopts a symmetric function such as max-pooling to collect these features into a global feature representation. Because the max-pooling layer only captures the maximum activation across global points, PointNet cannot learn to exploit local features. Building on PointNet, PointNet++~\cite{qi2017pointnet++} defines a hierarchical learning architecture. It hierarchically samples points using farthest point sampling (FPS) and groups local regions using k nearest neighbor search and ball search. Progressively, a simplified PointNet exploits features in local regions at multiple scales or resolutions. Similarly, Engelmann~et~al.~\cite{engelmann2018know} define local regions by KNN and K-means clustering and use a simplified PointNet to extract local features.

To learn the short- and long-range dependencies, some works introduce RNNs to MLP-based methods. For example, ESC~\cite{engelmann2017exploring} divides global points into multi-scale/grid blocks. The concatenated (local) block features are appended to the pointwise features and passed through Recurrent Consolidation Units (RCUs) to further learn global context features. Similarly, HRNN~\cite{ye20183d} uses pointwise pyramid pooling (3P) to extract local features from multi-size local regions. Pointwise features and local features are concatenated, and a two-direction hierarchical RNN explores context features on these concatenated features. However, the local features learned are insufficient because the deeper layer features do not cover a larger spatial extent.

Some works integrate the hand-crafted point representation into the PointNet or PointNet++ network to enhance the point representation ability with less learnable network parameters. Inspired by SIFT representation~\cite{lowe2004distinctive}, PointSIFT~\cite{jiang2018pointsift} inserts a PointSIFT module layer to learn local shape information. This module transforms each point into a new shape representation by encoding information about different orientations. PointWeb~\cite{zhao2019pointweb} proposes an adaptive feature adjustment (AFA) module to learn the interactive information between local points to enhance the point representation. Similarly, RepSurf~\cite{ran2022surface} introduces two novel point representations, namely triangular and umbrella representative surfaces, to establish connections and enhance the representation capability of learned point-wise features. 

This approach effectively improves feature representation with fewer learnable network parameters, drawing significant attention from the research community.
In contrast to the aforementioned methods, PointNeXt~\cite{qian2022pointnext} 
takes a different approach by revisiting the classical PointNet++ architecture through a systematic study of model training and scaling strategies. It proposes improved training strategies that lead to a significant performance boost for PointNet++. Additionally, PointNeXt 
introduces an inverted residual bottleneck design and employs separable MLPs to enable efficient and effective model scaling. Similarly, PointVector~\cite{deng2023pointvector} proposes a Vector-oriented point set abstraction that can aggregate neighboring features through high-dimensional vectors.

\vspace{2mm}\noindent\textbf{Point convolution techniques:} perform convolution operations directly on the points. Different from 2D convolution, the weight function of a point convolution needs to be learned from point geometric information adaptively. Early convolutional networks focused on the convolution weight function design. For example, RSNet~\cite{huang2018recurrent} exploits point-wise features using 1$\times$1 convolution and then passes them through the local dependency module (LDM) to exploit local context features. However, it does not define each point's neighborhood to learn local features. On the other hand, PointwiseCNN~\cite{hua2018pointwise} sorts points in a specific order, e.g., XYZ coordinates or Morton cureve~\cite{morton1966computer}, and queries nearest neighbors dynamically and bins them into 3$\times$3$\times$3 kernel cells before convolving with the same kernel weights.

Gradually, some point convolution methods approximate the convolution weight function as MLP to learn weights from point coordinates. PCCN~\cite{wang2018deep} performs parametric CNN, where the kernel is estimated as an MLP, on KD-tree neighborhoods to learn local features. PointCNN~\cite{li2018pointcnn} coarsens the input points with the FPS. The convolution layer learns an $\chi$ transformation from local points by MLP to simultaneously weight and permute the features, subsequently applying a standard convolution to these transformed features.

Some work associates a coefficient (derived from point coordinates) with the weight function to adjust the learned convolutional weights. An extension of the Monte Carlo approximation for convolution called PointConv~\cite{wu2019pointconv} considers the point density. It uses MLP to approximate a weight function of the convolution kernel and applies an inverse density scale to reweight the learned weight function. Similarly, MCC~\cite{hermosilla2018monte} phrases convolution as a Monte Carlo integration problem by relying on the joint probability density function (PDF), where an MLP also represents the convolution kernel. Moreover, it introduces Poisson-disk sampling (PDS)~\cite{wei2008parallel} to construct a point hierarchy instead of FPS, which provides an opportunity to get the maximal number of samples in a receptive field.

Another line of work employs another function instead of MLP to approximate the convolution weight function. Flex-Convolution~\cite{groh2018flex} uses a linear function with fewer parameters to model a convolution kernel and adapts inverse density importance sub-sampling (IDISS) to coarsen the points. KPConv~\cite{thomas2019kpconv} and KCNet~\cite{shen2018mining} fixed the convolution kernel for robustness to varying point densities. These networks predefine the kernel points on the local region and learn convolutional weights on the kernel points from their geometric connections to local points using linear and Gaussian correlation functions, respectively. Furthermore, KPConvX~\cite{thomas2024kpconvx} scales the depth wise convolutional weights with kernel attention values. Here, the number and position of kernel points need to be optimized for different datasets.

Point convolution on a limited local receptive field could not exploit long-range features. Therefore, some works introduce the dilated mechanism into point convolution. Dilated point convolution (DPC)~\cite{engelmann2020dilated} adapts standard point convolution to the neighborhood points of each point, where the neighborhood points are determined through a dilated KNN search. Similarly, A-CNN~\cite{komarichev2019cnn} defines a new local ring-shaped region by dilated KNN and projects points on a tangent plane to further order neighbor points in local regions. Then, the standard point convolutions are performed on these ordered neighbors, represented as a closed-loop array.

In the large-scale point cloud semantic segmentation area, RandLA-Net~\cite{hu2020randla} uses random point sampling instead of the more complex point selection approach. It introduces a novel local feature aggregation module (LFAM) to increase the receptive field and effectively preserve geometric details progressively. Another technology, PolarNet~\cite{zhang2020polarnet}, first partitions a large point cloud into smaller grids (local regions) along their polar bird’s-eye-view (BEV) coordinates. It then abstracts local region points into a fixed-length representation using a simplified PointNet, and these representations are passed through a standard convolution.

\vspace{2mm}\noindent\textbf{Graph convolution methods:} perform convolution on points connected with a graph structure, where the graph helps the feature aggregation exploit the structure information between points. The graphs can be divided into spectral graphs and spatial graphs. In the spectral graph, LS-GCN~\cite{wang2018local} adopts the basic architecture of PointNet++, replaces MLPs with a spectral graph convolution using standard ~unparameterized Fourier kernels, as well as a novel recursive spectral cluster pooling substitute for max-pooling. However, transformation from the spatial to the spectral domain incurs a high computational cost. Besides that, spectral graph networks are usually defined on a fixed graph structure and are thus unable to process data with varying graph structures directly.

In the spatial graph category, ECC~\cite{simonovsky2017dynamic} is among the pioneering methods to apply spatial graph networks to extract features from point clouds. It dynamically generates edge-conditioned filters to learn edge features describing relationships between a point and its neighbors. Based on PointNet architecture, DGCNN~\cite{wang2019dynamic} implements a dynamic edge convolution called EdgeConv in the neighborhood of each point. A simplified PointNet approximates the convolution. SPG~\cite{landrieu2018large} parts the point clouds into a number of simple geometrical shapes (termed super-points) and builds a super graph on global super-points. Furthermore, this network adopts PointNet to embed these points and refines the Gated Recurrent Unit (GRU) embedding. Based on the basic architecture of PoinNet++, Li~et~al.~\cite{li2019tgnet} proposed Geometric Graph Convolution (TGCov), with its filters defined as products of local point-wise features with local geometric connection features expressed by Gaussian weighted Taylor kernels. Feng~et~al.~\cite{feng2020point} constructed a local graph on neighborhood points searched along multi-directions and explored local features by a local attention-edge convolution (LAE-Conv). These features are imported into a point-wise spatial attention module to capture accurate and robust local geometric details. Lei~et~al.~\cite{lei2020seggcn} design a fuzzy coefficient to times weight function to enable the convolution weights to be robust.

Continuous graph convolution also incurs a high computational cost and generally suffers from the vanishing gradient problem. Inspired by the separable convolution strategy in Xception~\cite{chollet2017xception} that significantly reduces parameters and computation burden, HDGCN~\cite{liang2019hierarchical} designed a DGConv that composes depth-wise graph convolution followed by a point-wise convolution and adds DGConv to the hierarchical structure to extract local and global features. DeepGCNs~\cite{li2019deepgcns} borrow some concepts from 2D CNN, such as residual connections between different layers (ResNet), to alleviate the vanishing gradient problem and dilation mechanisms to allow the GCN to go deeper. The discrete spherical convolution kernel (SPH3D kernel)~\cite{lei2020spherical} consists of spherical convolution learning depth-wise features while point-wise convolution learning point-wise features.

Tree structures such as KD-tree and Octree can be viewed as a particular type of graph, allowing for the sharing of convolution layers depending on the tree splitting orientation. 3DContextNet~\cite{zeng20183dcontextnet} adopts a KD-tree structure to hierarchically represent points where the nodes of different tree layers represent local regions at various scales and employs a simplified PointNet with a gating function on nodes to explore local features. However, their performance depends heavily on the randomization of the tree construction. Lei~et~al.~\cite{lei2019octree} built an Octree-based hierarchical structure on global points to guide the spherical convolution computation per network layer. The spherical convolution kernel systematically partitions a 3D spherical region into multiple bins that specify learnable parameters to weight the points falling within the corresponding bin.

\vspace{2mm}\noindent\textbf{Transformer-based approaches:} have recently become famous for improving point cloud segmentation accuracy. Compared to point convolution, the Transformer introduces point features into weight learning. For example, the authors of~\cite{ma2020global} use the channel self-attention mechanism to learn independence between any two point-wise feature channels and further define a channel graph where the channel maps are presented as nodes and the independence as graph edges. AGCN~\cite{xie2020point} integrates the attention mechanism with GCN for analyzing the relationships between local features of points and introduces a global point graph to compensate for the relative information of individual points. Likewise, PointANSL~\cite{yan2020pointasnl} utilizes the general self-attention mechanism for group feature updating and proposes an adaptive sampling (AS) module to overcome the issues of FPS.

The transformer employs self-attention as a fundamental component and includes position encoding to capture the sequential order of input tokens. Position encoding is crucial to ensure that the model understands the relative positions of tokens within a sequence. Point Transformer~\cite{zhao2021point} introduces MLP-based position encoding into vector attention and uses a KNN-based downsampling module to decrease the point resolution. A follow-up work is Point Transformer v2~\cite{wu2022point}, strengthening the position encoding mechanism by applying an additional encoding multiplier to the relation vector and designing a partition-based pooling strategy to align the geometric information. FPTransformer~\cite{he2024full} introduces the full point encoding into the local point transformer to simultaneously learn the local, global and local-global features. 

Point transformers are typically computationally expensive because the original self-attention module needs to generate a considerable attention map. To address this problem, PatchFormer~\cite{zhang2022patchformer} calculates the attention map via low-rank approximation. Similarly, FastPointTransformer~\cite{park2022fast} introduces a lightweight local self-attention module that learns continuous positional information while reducing space complexity. Motivated by the success of the window-based transformer in the 2D domain, Stratified Transformer~\cite{lai2022stratified} designs a cubic window and samples distant points as keys, but more sparsely, to expand the receptive field. Besides, SphereFormer~\cite{lai2023spherical} designs radial window self-attention that partitions that space into several non-overlapping narrow and long windows for exploiting long-range dependencies. Most of transformer methods apply the transformer in a local region such as spherical or cubic window, which requires high computational cost. ConDaFormer~\cite{duan2024condaformer} address this problem by disassembling cubic windows into orthogonal 2D planes and enhancing local structures with depth wise convolution. To balance the accuracy and efficiency, Point Transformer v3~\cite{wu2024point} prioritizes simplicity and efficiency over the accuracy of certain mechanisms. This model replacing the precise neighbor search by KNN with an efficient serialized neighbor mapping, which expand the receptive field form 16 to 1024 points while remaining efficient.

\subsection{3D Video Based Segmentation}
Compared to the 3D single frame/scan semantic segmentation methods reviewed earlier, 3D video (continuous frames/scans) semantic segmentation methods take into account the connecting spatiotemporal information between frames, which is more effective at parsing the scene robustly and continuously. Conventional CNNs are not designed to exploit the temporal information between frames. A common strategy is to adapt recurrent neural networks or spatiotemporal convolutional networks.

\vspace{2mm}\noindent\textbf{RNNs:}~generally work in combination with 2D CNNs to process RGB-D videos. The 2D CNN extracts the frame-wise spatial information, and the RNN learns the temporal information between the frames. Valipour~et~al.~\cite{valipour2017recurrent} proposed a recurrent fully neural network to operate over a sliding window over the RGB-D video frames. Specifically, the convolutional gated recurrent unit preserves the spatial information and reduces the parameters. Similarly, In~\cite{emre2017semantic}, Yurdakul~et~al. combine a fully convolutional and RNN to investigate the contribution of depth and temporal information separately in the synthetic RGB-D video.

\vspace{2mm}\noindent\textbf{Spatio-temporal convolution methods:} Nearby video frames provide diverse viewpoints and additional context for objects and scenes. STD2P~\cite{he2017std2p} operates a novel spatiotemporal pooling layer to aggregate region correspondences computed by optical flow and image boundary-based super-pixels. Choy~et~al.~\cite{choy20194d} proposed 4D Spatio-Temporary ConvNet to process a 3D point cloud video directly. To overcome challenges in high-dimensional 4D space (3D space and time), they introduced the 4D spatial-temporal convolution, a generalized sparse convolution, and the trilateral-stationary conditional random field that keeps spatiotemporal consistency. Likewise, based on 3D sparse convolution, SpSequenceNet~\cite{shi2020spsequencenet} contains two novel modules, a cross-frame global attention module and a cross-frame local interpolation module, to exploit spatial and temporal features in 4D point clouds. PointMotionNet~\cite{wang2022pointmotionnet} proposes a spatio-temporal convolution that exploits a time-invariant spatial neighboring space and extracts spatiotemporal features to distinguish between moving and static objects. TVSN~\cite{shi2024learning} built a temporal graph on the 3D point cloud sequences and captures the temporal variation with a graph convolution, transforming coarse predictions into fine predictions.

\vspace{2mm}\noindent\textbf{Spatio-temporal transformer techniques:} Point tracking is usually employed to capture the dynamics in point cloud videos. However, P4Transformer~\cite{fan2021point} offers a 4D convolution to embed the spatiotemporal local structures in point cloud video and further introduces a transformer to leverage the motion information across the entire video by performing the self-attention on these embedded local features. Also, PST$^2$~\cite{wei2022spatial} performs spatiotemporal self-attention across adjacent frames to capture the spatiotemporal context and proposes a resolution embedding module to enhance the resolution of feature maps by aggregating features. X4D-Transformer~\cite{jing2024x4d} leverages texture priors from RGB sequences using a dual-branch transformer into 3D point clouds sequences, to enhance 3D video understanding.

\subsection{Other Representation Based Methods}
Some methods transform the original point cloud into representations other than projected images, voxels, and points. Examples of such representations include tangent images~\cite{tatarchenko2018tangent} and lattice~\cite{su2018splatnet},~\cite{rosu2019latticenet}. In the former case, Tatarchenko~et~al.~\cite{tatarchenko2018tangent} project local surfaces around each point to a series of 2D tangent images and develop a tangent convolution-based U-Net to extract features. In the latter case, SPLATNet~\cite{su2018splatnet} adapts the bilateral convolution layers (BCLs) proposed by Jampani~et~al.~\cite{jampani2016learning} to map disordered points onto a sparse lattice smoothly. Besides, LatticeNet~\cite{rosu2019latticenet} employs a hybrid architecture that combines PointNet, which obtains low-level features, with sparse 3D convolution, which explores global context features. These features are embedded into a sparse lattice that allows the application of standard 2D convolutions.

Although the above methods have achieved significant progress in 3D semantic segmentation, each has drawbacks. For instance, multi-view images have more spectral information, like color or intensity, but fewer geometric details on the scene. On the other hand, voxels have more geometric information but less spectral information. To get the best of both worlds, some methods adopt hybrid representations as input to learn comprehensive features of a scene. Dai~et~al.~\cite{dai20183dmv} map 2D semantic features obtained by multi-view networks into 3D grids of scenes. These pipelines make 3D grids attach rich 2D semantic as well as 3D geometric information so that the scene can get better segmentation by a 3D CNN.

Likewise, Hung~et~al.~\cite{chiang2019unified} back-project 2D multi-view image features onto the 3D point cloud space and use a unified network to extract local details and global context from sub-volumes and the global scene, respectively. Liu~et~al.~\cite{liu2019point} argue that voxel-based and point-based NN are computationally inefficient in high-resolution and data structuring. To overcome these challenges, they propose Point-Voxel CNN (PVCNN), which represents the 3D input data as point clouds to take advantage of the sparsity to lower the memory footprint and leverage the voxel-based convolution to obtain a contiguous memory access pattern. Jaritz~et~al.~\cite{jaritz2019multi} proposed MVPNet that collects 2D multi-view dense image features into 3D sparse point clouds and then uses a unified network to fuse the semantic and geometric features. Also, Meyer~et~al.~\cite{meyer2019sensor} fuse 2D image and point clouds to address 3D object detection and semantic segmentation by a unifying network. BPNet~\cite{hu2021bidirectional} consists of 2D and 3D sub-networks with symmetric architectures connected through a bidirectional projection module (BPM). This allows the interaction of complementary information from both visual domains at multiple architectural levels, improving scene recognition by leveraging the advantages of both 2D and 3D information. The other representation-based semantic segmentation methods are summarized in Table~\ref{table3}.

\section{3D Instance Segmentation}\label{section4}
3D instance segmentation methods additionally distinguish between different instances of the same class. Being a more informative task for scene understanding, 3D instance segmentation is receiving increased interest from the research community. 3D instance segmentation methods are roughly divided into two directions: proposal-based and proposal-free.

\begin{figure}[tbp]
\centering
\includegraphics[width=0.95\columnwidth]{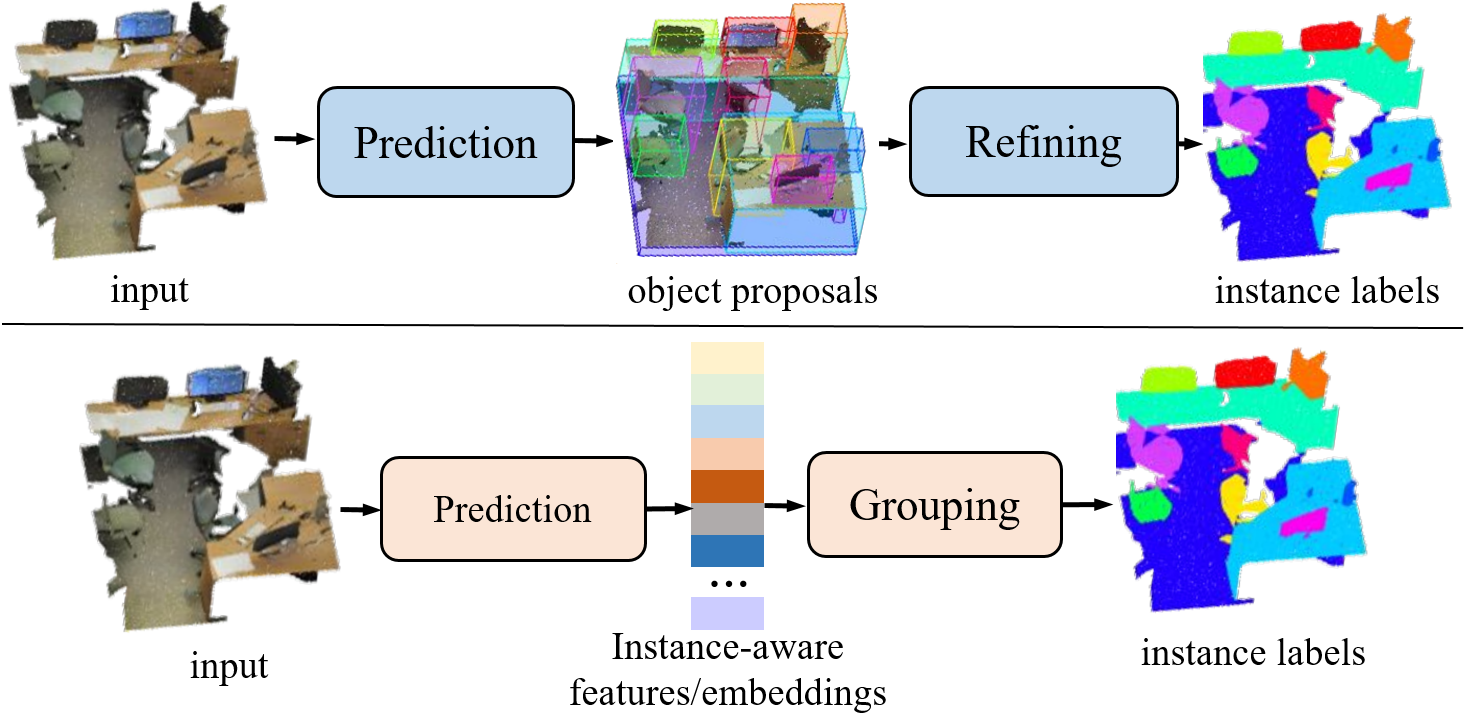}
\caption{Illustration of two different approaches for 3D instance segmentation. \textbf{Top row}: proposal-based framework. \textbf{Bottom row}: proposal-free framework.}
\vspace{-2mm}
\label{instancesegmentation}
\end{figure}

\subsection{Proposal-Based Methods}

Proposal-based methods first predict object proposals and then refine them to generate final instance masks (see Figure~\ref{instancesegmentation}), breaking down the task into two main challenges. Hence, from the proposal generation point of view, these methods can be grouped into detection-based and detection-free methods.

\vspace{2mm}\noindent\textbf{Detection-based methods:} often \textcolor{black}{treat} object proposals as a 3D bounding box regression problem. \textcolor{black}{For example}, 3D-SIS~\cite{hou20193d} integrates high-resolution RGB images with voxels, \textcolor{black}{aligned according to the 3D reconstruction pose,} and jointly learns color and geometric features \textcolor{black}{using a 3D detection backbone} to predict 3D bounding box proposals. In these proposals, a 3D mask backbone is used to predict the final instance masks. Similarly, GPSN~\cite{yi2019gspn} introduces a 3D object proposal network that reconstructs object shapes from noisy observations to improve geometric understanding. \textcolor{black}{GPSN is also embedded into a 3D instance segmentation network called Region-based PointNet (R-PointNet), which refines, accepts, or rejects proposals. These networks require a step-by-step training process, and proposal refinement can be computationally expensive due to suppression operations. To overcome these challenges,} 3D-BoNet~\cite{yang2019learning} proposes an innovative end-to-end network that directly learns a fixed number of 3D bounding boxes \textcolor{black}{without needing proposal rejection, followed by the estimation of} an instance mask within each bounding box.

\begin{table*}[t]
\centering
\caption{Summary of 3D instance segmentation methods with deep learning. M$\leftarrow$multi-view image; Me$\leftarrow$mesh;V$\leftarrow$voxel;  P$\leftarrow$point clouds. }
\label{table5}
\scriptsize
\centering
\renewcommand\arraystretch{1.4}
\setlength{\tabcolsep}{2mm}{\begin{tabular*}{\textwidth}{@{\extracolsep\fill}l|lcll|l}
\specialrule{1pt}{0pt}{0pt} 
& \textbf{Methods}    & \textbf{Input} & \textbf{Propo./Embed. Prediction}  & \textbf{Refining/Grouping}      & \textbf{Contribution}       \\\hline
\multirow{8}{*}{\begin{sideways}Proposal based\end{sideways}}
                                & GSPN~\cite{yi2019gspn}        & P     & GSPN                                   & R-PointNet                   & New proposal generation methods  \\
                                & 3D-SIS~\cite{hou20193d}      & M+V & 3D-RPN+3D-RoI                          & 3DFCN                        & Learning bounding box on geometry and RGB \\
                                & 3D-BoNet~\cite{yang2019learning}    & P     & Bounding box regression             & Point mask prediction           & Directly regressing 3D bounding box     \\
                                & SGPN~\cite{wang2018sgpn}       & P     & SM + SCM + PN & Non-Maximum suppre.         & New group proposal           \\
                                & 3D-MPA~\cite{engelmann20203d}      & P     & SSCNet                          & Graph ConvNet                       & Multi proposal aggregation strategy                 \\
                                & AS-Net~\cite{jiang2020end}      & P     & Four branches with MLPs        & Candidate prop. suppre.       & Novel Algorithm mapping labels to candidates    \\
                                
                                & SoftGroup~\cite{vu2022softgroup}   & P     & Soft-grouping module    & top-down refinment   & Novel clustering algorithm based on dual coordinate sets \\ 
                                
                                & SSTNet~\cite{liang2021instance}      & p     & Tree traversal + splitting        & CliqueNet        & Constructing the superpoint tree for instance segmentation    \\
                                
                                \hline
\multirow{15}{*}{\begin{sideways}Proposal free\end{sideways}}
                                & 3D-BEVIS~\cite{elich20193d}    & M     & U-Net/FCN + 3D prop.   & Mean-shift clustering & Joint 2D-3D feature                                  \\
                                & PanopticFus~\cite{narita2019panopticfusion} & M     & PSPNet/Mask R-CNN    & FC-CRF                        & Coopering with semantic mapping                                                                             \\
                                & ASIS~\cite{wang2019associatively}       & P     & 1 encoder+ 2 decoders     & ASIS module        & Simultaneously performing sem./ins.   segmentation tasks      \\
                                & JSIS3D~\cite{pham2019jsis3d}      & P     & MT-PNet               & MV-CRF                       & Simultaneously performing sem./ins.   segmentation tasks                                                    \\
                                & 3D-GEL~\cite{liang20193d}      & P     & 3D U-Net               & GCN                          & Structure-aware loss function +   attention-based GCN                                                       \\
                                & OccuSeg~\cite{han2020occuseg}     & P     & 3D U-Net            & Graph-based  clustering     & Proposing a novel occupancy signal                \\
                                
                                & MASC~\cite{liu2019masc}        & Me    & 3D U-Net    & Clustering algorithm          & Novel clustering based on affinity and mesh topology                                         \\
                                
                                & MTML~\cite{lahoud20193d}       & V     & 3D U-Net               & Mean-shift clustering         & Multi-task learning \\

                                & PointGroup~\cite{jiang2020pointgroup}   & P     & 3D U-Net    & Clustering + ScoreNet   & Novel clustering algorithm based on dual coordinate sets \\   

                                & HAIS~\cite{chen2021hierarchical}   & P     & 3D U-Net    & Set aggregation   & Hierarchical aggregation for fine-grained predictions \\  

                                & Dyco3D~\cite{he2021dyco3d}   & P     & 3D U-Net    & Dynamic conv. & Generating kernel by clustering for convolution  \\  

                                & PointInst3D~\cite{he2022pointinst3d}   & P     & 3D U-Net    & MLP   & Generating kernel by FPS \\  

                                & DKNet~\cite{wu20223d}   & P     & 3D U-Net    & MLP   & Generating kernel by candidate mining and aggregation \\  
                                
                                & ISBNet~\cite{ngo2023isbnet}   & P     & 3D U-Net    & Box-aware dynamic conv   & Generating kernel by instance aware FPS and point aggrega. \\ 
                                
                                & Spherical Mask ~\cite{shin2024spherical} &P     & 3D U-Net    & Rradial point migration   & Radial instance detection \\
                                
                                & SPFormer ~\cite{sun2023superpoint}   &P     & 3D U-Net    & Query Decoder   & Generating instance by transformer \\
                                
                                & Mask3D ~\cite{schult2023mask3d}  &P     & 3D U-Net    & Transformer Decoder   & Generating instance by transformer \\

                                & QueryFormer ~\cite{lu2023query}  &P     & 3D U-Net    & Affiliated Transformer 
                                Decoder   & Generating instance by transformer \\

                                &Oneformer~\cite{kolodiazhnyi2024oneformer3d}   &P     & 3D U-Net    & Transformer Decoder   & A unify network for 3D segmentation \\
                              
                               \specialrule{1pt}{0pt}{0pt} 

\end{tabular*}}
\end{table*}


\vspace{2mm}\noindent\textbf{Detection-free methods:}, include SGPN~\cite{wang2018sgpn}, which assumes that points belonging to the same object instance should have similar features. It learns a similarity matrix to predict proposals, and confidence scores of the points are used to prune proposals, generating highly reliable instance predictions. However, this simple distance-based similarity metric lacks informativeness and struggles to segment adjacent objects of the same class. \textcolor{black}{In contrast, 3D-MPA~\cite{engelmann20203d} generates object proposals by learning from sampled and grouped point features that vote for the same object center, then refines the proposal features using a graph convolutional network, enabling higher-level interactions between proposals for more accurate results.} AS-Net~\cite{jiang2020end} employs an assignment module to allocate proposal candidates and removes redundant ones through a suppression network. SoftGroup~\cite{vu2022softgroup} introduces top-down refinement for instance proposals. SSTNet~\cite{liang2021instance} presents an end-to-end Semantic Superpoint Tree Network (SSTNet), which generates object instance proposals directly from scene points. A key innovation of SSTNet is the construction of an intermediate semantic superpoint tree (SST) based on the learned semantic features of superpoints.

\subsection{Proposal Free Methods}

Proposal-free methods learn feature embeddings for each point and then apply clustering to obtain definitive 3D instance labels as shown in Figure~\ref{instancesegmentation}, breaking down the task into two main challenges. From the embedding learning point of view, these methods can be roughly subdivided into five categories: 2D embedding-based, multi-task learning, clustering-based, dynamic convolution-based and dynamic transformer-based.


\vspace{2mm}\noindent \textbf{2D embedding based strategies:}~One example is 3D-BEVIS~\cite{elich20193d}, which learns a 2D global instance embedding \textcolor{black}{from a bird’s-eye view of the entire scene and then propagates this embedding onto point clouds using DGCNN~\cite{wang2019dynamic}. Another example is PanopticFusion~\cite{narita2019panopticfusion}, which employs the 2D instance segmentation network Mask R-CNN~\cite{he2017mask} to predict pixel-wise instance labels for RGB frames and then incorporates these predicted labels into 3D volumes}.


\vspace{2mm}\noindent\textbf{Multi-tasks learning methods:}, such as 3D semantic and 3D instance segmentation, can mutually enhance each other. For example, \textcolor{black}{objects from different classes must be distinguished as separate instances, while those sharing the same instance label should belong to the same class. Building on this concept}, ASIS~[\textcolor{cyan}{181}] \textcolor{black}{introduces an encoder-decoder network designed to learn semantic-aware instance embeddings, improving the performance of both tasks.} Similarly, JSIS3D~[\textcolor{cyan}{134}] utilizes a unified network, named MT-PNet, to predict semantic labels for points and embed them into high-dimensional feature vectors, \textcolor{black}{while also employing MV-CRF to jointly optimize object classes and instance labels. Likewise,} Liu et al.~[\textcolor{cyan}{108}] and 3D-GEL~[\textcolor{cyan}{106}] leverage SSCN to simultaneously produce semantic predictions and instance embeddings, with two GCNs refining the instance labels. OccuSeg~[\textcolor{cyan}{47}] applies a multi-task learning network to generate both occupancy signals and spatial embeddings, where the occupancy signal reflects the number of voxels occupied per voxel.


\vspace{2mm}\noindent\textbf{Clustering based techniques:} like MASC~\cite{liu2019masc}, \textcolor{black}{leverage the powerful capabilities of SSCN~\cite{graham20183d}} to predict similarity embeddings between neighboring points across multiple scales and semantic structures. \textcolor{black}{A straightforward yet effective} clustering technique~\cite{liu2018affinity} is applied to segment points into instances \textcolor{black}{using these two types of learned embeddings.} MTML~\cite{lahoud20193d} employs two sets of feature embeddings: \textcolor{black}{one for capturing unique instance-specific features and another for direction embedding}, providing stronger cohesion for grouping. Similarly, PointGroup~\cite{jiang2020pointgroup} \textcolor{black}{forms clusters by combining original and shifted coordinate embedding spaces, with ScoreNet assisting in selecting optimal clusters.} While these approaches group points based on point-level embeddings, they lack instance-level adjustments. HAIS~\cite{chen2021hierarchical} addresses this by introducing set aggregation and intra-instance prediction to refine object-level instances.

\vspace{2mm}\noindent\textbf{Dynamic convolution-based approaches:} overcome the limitations of clustering-based methods by generating kernels and then convolving with the point features to generate instance masks. Dyco3D~\cite{he2021dyco3d} adopts the clustering algorithm to generate a kernel for convolution. Similarly, PointInst3D~\cite{he2022pointinst3d} uses FPS to generate kernels. DKNet~\cite{wu20223d} introduces candidate mining and candidate aggregation to generate more instance kernels. Moreover, ISBNet~\cite{ngo2023isbnet} proposes a new instance encoder combining instance-aware PFS with a point aggregation layer to generate kernels to replace clustering in DyCo3D. Spherical Mask~\cite{shin2024spherical} uses the similar backbone as ISBNet. It introduce spherical representation to overcome size overestimation and error propagation, significantly improving instance segmentation performance.

\vspace{2mm}\noindent\textbf{Dynamic transformer-based approaches:} can learn more precious context information, being more conducive to identify the instances. SPFormer~\cite{sun2023superpoint} groups potential features from point clouds into superpoint and predicts instance through query attention without relying on the results of detection and segmentation. Mask3D~\cite{schult2023mask3d} directly predict instance mask from point clouds using instance queries, eliminating the need for traditional voting mechanisms and geometric clustering techniques. Similarly, QueryFormer~\cite{lu2023query} enhances instance segmentation by optimizing query initialization for better coverage and reducing noise through a specialized transformer decoder, resulting in more accurate instance masks and semantic labels.  Oneformer~\cite{kolodiazhnyi2024oneformer3d} introduces a unified 3D segmentation framework performs 3D segmentation  using a
group of learnable kernels. These kernels are trained with a transformer based decoder with unified instance and semantic queries
passed as an input.   Table~\ref{table5} summarizes 3D instance segmentation methods.


\section{3D Part Segmentation}\label{section5}
3D part segmentation is the next finer level, after instance segmentation, where the aim is to label different parts of an instance. The pipeline of part segmentation is quite similar to semantic segmentation, except that the labels are now for individual parts. Therefore, some existing 3D semantic segmentation networks \cite{meng2019vv}, \cite{graham20183d}, \cite{qi2017pointnet}, \cite{qi2017pointnet++}, \cite{zeng20183dcontextnet}, \cite{huang2018recurrent}, \cite{thomas2019kpconv}, \cite{hua2018pointwise}, \cite{hermosilla2018monte}, \cite{wu2019pointconv}, \cite{li2018pointcnn}, \cite{wang2019dynamic}, \cite{lei2020spherical}, \cite{xie2020point}, \cite{wang2018deep}, \cite{groh2018flex}, \cite{lei2019octree}, \cite{su2018splatnet}, \cite{rosu2019latticenet}, \cite{deng2023pointvector}, \cite{qian2022pointnext}, \cite{zhao2021point}can also be trained for part segmentation. However, these networks can not entirely tackle the difficulties of part segmentation. For example, various parts with the same semantic label might have diverse shapes, and the number of parts for an instance with the same semantic label may be different. We subdivide 3D part segmentation methods into the following categories: regular data based and irregular data based.

\subsection{Segmentation of Regular Data}
Regular data usually includes projected images~\cite{kalogerakis20173d}, voxels~\cite{wang2019voxsegnet},~\cite{le2018pointgrid},~\cite{song2017embedding}. As for projected images, Kalogerakis~et~al.~\cite{kalogerakis20173d} obtain a set of images from multiple views that optimally cover the object surface and then use multi-view Fully Convolutional Networks (FCNs) and surface-based Conditional Random Fields (CRFs) to predict and refine part labels separately. Voxel is a useful representation of geometric data. However, fine-grained tasks like part segmentation require high-resolution voxels with more detailed structure information, which leads to high computation costs. VoxSegNet~\cite{wang2019voxsegnet} exploits more detailed information from voxels with limited resolution. They use spatial dense extraction to preserve the spatial resolution during the sub-sampling process and an attention feature aggregation (AFA) module to adaptively select scale features. PointGrid~\cite{le2018pointgrid} incorporates a constant number of points with each cell, allowing the network to learn better local geometry shape details. Furthermore, multiple model fusion can enhance segmentation performance. Combining the advantages of images and voxels, Song~et~al.~\cite{song2017embedding} proposed a two-stream FCN, termed AppNet and GeoNet, to explore 2D appearance and 3D geometric features from 2D images. In particular, their VolNet extracts 3D geometric features from 3D volumes, guiding GeoNet in extracting features from a single image.

\begin{table*}[tbp]
\caption{Summary of 3D part segmentation methods. M$\leftarrow$multi-view image; Me$\leftarrow$mesh; V$\leftarrow$voxel; P$\leftarrow$point clouds; reg.$\leftarrow$regular data; irreg.$\leftarrow$irregular data; MS.$\leftarrow$Multi-Stream; 2S.$\leftarrow$2-Stream.}
\label{table6}
\renewcommand\arraystretch{1.1}
\centering
\setlength{\tabcolsep}{0.5mm}{
\begin{tabular*}{\textwidth}{c|l|c|c|c|l}
\specialrule{1pt}{0pt}{0pt} 
& \textbf{Methods}      & \textbf{Input} & \textbf{Archi.}             & \textbf{Backbone}       & \textbf{Contribution}                                                                                                                          \\\hline
\multirow{4}{*}{\begin{sideways}Regular\end{sideways}}   & ShapePFCN~\cite{kalogerakis20173d}      & M     & MS-FCN       & 2DConv                  & Per-label confidence maps +   surface-based CRF  \\
                                 & VoxSegNet~\cite{wang2019voxsegnet}      & V     & 3DU-Net                  & AtrousConv              & SDE for  preserving the spatial resolution AFA for feature selecting \\
                                 & Pointgrid~\cite{le2018pointgrid}      & V     & Conv-deconv              & 3DConv                  & Learning higher order local geometry   shape.                                                                                         \\
                                 & SubvolumeSup~\cite{song2017embedding}  & M+V   & 2S-FCN             & 2D/3DConv               & GeoNet/AppNet for 3/2D features exploi. + DCT for aligning.                                              \\\hline
\multirow{12}{*}{\begin{sideways}Irregular\end{sideways}} & DCN~\cite{xu2017directionally}            & Me    & 2S-DCN \& NN & DConv         & DCN/NN for local feature and global   feature.                                                                                        \\
                                 & MeshCNN~\cite{hanocka2019meshcnn}        & Me    & 2D CNN                    & MeshConv                & Novel mesh convolution and pooling                                                                                                    \\
                                 & PartNet~\cite{yu2019partnet}       & P     & RNN                      & PN                      & Part feature learning scheme for   context and geometry feature exploitation                                                          \\
                                 & SSCNN~\cite{yi2017syncspeccnn}    & P     & FCN                      & SpectralConv            & STN for allowing weight sharing,spectral   multi-scale kernel                                                                       \\

                                & \textcolor{black}{CurveNet~\cite{xiang2021walk}}    & \textcolor{black}{P}     & \textcolor{black}{PN++}                      &\textcolor{black}{ MLP}            & \textcolor{black}{A hypothetical curve representation}\\

                                &\textcolor{black}{ PointMLP~\cite{marethinking}}    & \textcolor{black}{P}     & \textcolor{black}{PN++}                      & \textcolor{black}{MLP}            & \textcolor{black}{A lightweight geometric affine module}\\
                                
                                & \textcolor{black}{SPoTr~\cite{park2023self}}    & \textcolor{black}{P}     & \textcolor{black}{PN++ }                     & \textcolor{black}{Transformer}            &\textcolor{black}{Self-positioning global cross-attention}\\
              
                                 & KCNet~\cite{shen2018mining}          & P     & PN                       & MLP                     & KNN graph on points, kernel   correlation for measuring geometric affinity                                                           \\
                                 & SFCN~\cite{wang20183d}          & P     & FCN                      & SFConv                  & Novel point convolution                                                                                                               \\
                                 & SpiderCNN~\cite{xu2018spidercnn}      & P     & PN                       & SpiderConv              & Novel point   convolution                                                                                                             \\
                                 & FeaStNet~\cite{verma2018feastnet}       & P     & U-Net                    & GConv                   & Dynamic graph convolution filters                                                                                                     \\
                                 & Kd-Net~\cite{klokov2017escape}       & P     & Kd-tree                  & Affine Transfor.   & Useing Kd-tree to build graphs and share learnable parameters                         \\
                                 & O-CNN~\cite{wang2017cnn}          & P     & Octree                   & 3DConv                  & Making 3D-CNN feasible for high-resolu.   voxels                                                                                       \\
                                 & PointCapsNet~\cite{zhao20193d}& P     & Enco.-deco.          & PN                      & Semi-supervision learning                                                                                                             \\
                                 & SO-Net~\cite{li2018so}        & P     & Enco.-deco.          & FC layers               & SOM for modeling spatial   distribution +  un-supervision  learning\\\specialrule{1pt}{0pt}{0pt} 
\end{tabular*}}
\end{table*}

\subsection{Segmentation of Irregular Data}
Irregular data representations usually include meshes \cite{xu2017directionally}, \cite{hanocka2019meshcnn} and point clouds \cite{li2018so}, \cite{shen2018mining}, \cite{yi2017syncspeccnn}, \cite{verma2018feastnet}, \cite{wang20183d}, \cite{yu2019partnet}, \cite{zhao20193d}, \cite{yue2022drgcnn}. Mesh provides an efficient approximation to a 3D shape because it captures the flat, sharp, and intricate surface shape and topology. Xu~et~al.~\cite{xu2017directionally} put the face normal and face distance histogram as the input of a two-stream framework and use the CRF to optimize the final labels. Inspired by traditional CNN, Hanocka~et~al.~\cite{hanocka2019meshcnn} design novel mesh convolution and pooling to operate on the mesh edges.

\textcolor{black}{Similar to MLP-based methods for point clouds semantic segmentation, CurveNet~\cite{xiang2021walk} aggregates hypothetical curves in point clouds to enhance point feature learning. PointMLP \cite{marethinking} introduces a lightweight geometric affine module to capture diverse geometric structures across different local regions. In the transformer domain, SPoTr~\cite{park2023self} combines local self-attention and self-positioning global cross-attention to learn local and global features. Although Point Transformer has strong context-learning capabilities, its large number of parameters makes it challenging to achieve good performance on limited datasets. }

Graph convolution is the most commonly used pipeline. SyncSpecCNN~\cite{yi2017syncspeccnn} introduces a synchronized spectral CNN to process irregular data in the spectral graph domain. In particular, multichannel and parametrized dilated convolution kernels are proposed to solve multi-scale analysis and information sharing across shapes, respectively. In the spatial graph domain, in analogy to a convolution kernel for images, KCNet~\cite{shen2018mining} presents a point-set kernel and nearest-neighbor-graph to improve PointNet with an efficient local feature exploitation structure. Similarly, Wang~et~al.~\cite{wang20183d} design shape fully convolutional networks (SFCN) based on graph convolution and pooling operations, similar to FCN on images. SpiderCNN~\cite{xu2018spidercnn} applies a particular family of convolutional filters that combine simple step functions with Taylor polynomials, making the filters capture intricate local geometric variations effectively. Furthermore, FeastNet~\cite{verma2018feastnet} uses a dynamic graph convolution operator to build relationships between filter weights and graph neighborhoods instead of relying on the static graph of the above network.

A special kind of graph, the trees (e.g., Kd-tree and Octree), works on 3D shapes with different representations and can support various CNN architectures. Kd-Net~\cite{klokov2017escape} uses a kd-tree data structure to represent point cloud connectivity. However, the networks have high computational costs. O-CNN~\cite{wang2017cnn} designs an Octree data structure from 3D shapes. However, the computational cost of the O-CNN grows quadratically as the depth of the tree increases.

SO-Net~\cite{li2018so} sets up a self-organization map (SOM) from point clouds and hierarchically learns node-wise features on this map using the PointNet architecture. However, it fails to exploit local features fully. PartNet~\cite{yu2019partnet} decomposes 3D shapes top-down and proposes a Recursive Neural Network (RvNN) for learning the hierarchy of fine-grained parts. Zhao~et~al.~\cite{zhao20193d} introduce an encoder-decoder network, 3D-PointCapsNet, to tackle several common point cloud-related tasks. The dynamic routing scheme and the peculiar 2D latent space deployed by capsule networks in their model bring improved performance. The 3D part segmentation methods are summarized in Table~\ref{table6}.

\section{ Applications}

Deep learning based 3D segmentation technology holds significant value in cultural heritage preservation and semantic map construction.

\subsection{Cultural Heritage Preservation}

Cultural heritage preservation demands precise documentation of artifacts' shapes and textures for digital archiving, damage assessment, and research. Traditional methods of manual measurement and recording are often time-consuming and prone to errors. In contrast, deep learning based 3D segmentation provides an efficient and accurate solution for capturing and analyzing detailed artifact information. Deep learning based 3D segmentation in cultural heritage preservation can generally be divided into the surface/object-level, and scene-level segmentation.

\noindent\textbf{Surface/object-level segmentation:} targets the segmentation of different surface types or textures, like walls, floors, or detailed carvings, to provide a detailed representation of structural features. Ganapathi et al.~\cite{ganapathi2022detecting} develop a direct 3D mesh-based texture segmentation approach that identifies and classifies surface regions into texture and non-texture areas using a hybrid method combining classical features and a deep transformer. Furthermore, this team~\cite{ganapathi2023facet}  introduces a binary classification framework that classifies texture and non-texture regions on 3D surfaces at the facet level using a deep vision transformer, with input generated from local geometric properties, demonstrating effectiveness on diverse texture pattern datasets. Ji et al.~\cite{ji2023semantic} develop  a semantic segmentation and visualization framework that enhances the extraction of soft edges in cultural heritage reliefs using a deep learning based network, demonstrated on the Borobudur Temple bas-reliefs. They employ a novel opacity-based edge highlighting technique to extract soft edges, which are characteristic of reliefs, and incorporates them as guide information in a multichannel image input (including RGB, depth maps, and soft-edge images) to a custom network. Fu et al.~\cite{fu2024detecting} develop a Swin Transformer- and YOLOv5-based model for the automated classification, segmentation, and detection of surface defects in heritage buildings.

\noindent\textbf{Scene-level segmentation:} involves segmenting entire scenes or environments to understand the spatial context and layout of cultural heritage sites, capturing relationships between objects and features. Matrone et al.~\cite{matrone2020comparing} make a comparative analysis of machine learning and deep learning methods for semantic segmentation of large 3D cultural heritage point clouds, and propose a hybrid architecture named DGCNN-Mod+3Dfeat, to enhance complex scene recognition. Pierdicca et al.~\cite{pierdicca2020point} develop an enhanced deep learning framework for 3D point cloud segmentation in the Digital Cultural Heritage domain, utilizing an improved DGCNN that incorporates additional features like normal and color. Haznedar et al.~\cite{haznedar2023implementing} evaluate PointNet-based segmentation for heritage buildings, revealing its limitations with deformed and deteriorated structures, and propose a novel workflow that enhances segmentation accuracy by training PointNet with restitution-based synthetic data. Artopoulos et al.~\cite{artopoulos2023artificial} integrate deep neural networks (DNNs) and support vector machines (SVMs) for identifying architectural stylistic influences in Cypriot historical architecture. Zhao et al.~\cite{zhao2024dsc} propose a DSC-Net that leverages Enhanced Dual Attention Pooling and Global Context Feature Aggregation for precise 3D point cloud segmentation. Yang et al.~\cite{yang2024point} propose a novel network GSS-Net for point cloud semantic segmentation tailored for grotto scenes. This model incorporates knowledge guidance such as point cloud density, visual color, local geometric features, and global spatial distribution, to enhance segmentation accuracy.

\subsection{3D Semantic Map Construction}
Unmanned systems do not just need to avoid obstacles but also need to establish a deeper understanding of the scene such as object parsing, self localization etc. To facilitate such tasks, unmanned systems build a 3D semantic map of the scene which includes two key problems: geometric reconstruction and semantic segmentation. 3D scene reconstruction has conventionally relied on simultaneous localization and mapping system (SLAM) to obtain a 3D map without semantic information. This is followed by semantic segmentation. 3D semantic map construction can be appropriate to categorize into non-real-time approaches and real-time approaches.

\noindent \textbf{Non-real-time approaches:}~may be more suited for post-processing or tasks requiring high precision analysis. To obtain more information, some works learn the connections between frames to enhance the segmentation accuracy. For example, Xiang et al.~\cite{xiang2017rnn} propose a data associated recurrent neural networks (DA-RNN) and integrate the output of the DA-RNN, which provides a consistent semantic labeling of the 3D scene. Cheng et al.~\cite{cheng2020robust} use a CRF-RNN-based semantic segmentation to generate the corresponding labels. Similarly, Kochanov et al.  \cite{kochanov2016scene} also use scene flow to propagate dynamic objects within the 3D semantic maps. Some works fuses the multi-modal data to improve the segmentation accuracy. For example, Berrio et al.~\cite{li2020building} integrate LIDAR and camera data to build a 3D semantic voxelized map, incorporating uncertainties in sensor readings and enhancing semantic segmentation for autonomous vehicle navigation. Shi et al.~\cite{berrio2021camera} improve 3D semantic map generation from RGB-D scans by focusing on accurate 2D frame labeling and combining them in 3D space using a novel semantic fusion mechanism, enhanced by a two-stream network and discriminatory mask loss for robust semantic segmentation. 

\noindent \textbf{Real-time approaches:}~are typically used in tasks requiring immediate environmental awareness, such as autonomous driving and robotic navigation. Qin et al.~\cite{qin2021light} introduce a cost-effective localization framework for autonomous driving that utilizes low-cost cameras and compact visual semantic maps, incorporating semantic segmentation to accurately identify road elements for improved localization. Wilson et al.~\cite{wilson2022motionsc} develop a MotionSC algorithm to leverages 3D deep learning to enhance SSC with temporal information, improving real-time dense local semantic mapping in dynamic environments. Wang et al.~\cite{wang2023semlaps} introduce a real-time semantic mapping methodology that combines 2D and 3D networks within a SLAM system, enhancing image segmentation and 3D map processing, achieving state-of-the-art semantic mapping quality and superior cross-sensor generalization in real-time applications. Yamazaki et al.~\cite{yamazaki2024open} develop a real-time open-vocabulary 3D mapping approach that uses a vision-language foundation model and TSDF for swift 3D scene reconstruction, achieving annotation-free 3D segmentation and superior performance in open-world semantics without additional 3D training.

\vspace{2mm}
\section{Experimental Results}\label{section7}
Below, we summarize the quantitative results of the segmentation methods discussed earlier on some typical public datasets and analyze these results qualitatively.

\subsection{Results for 3D Semantic Segmentation}
We report the results of RGB-D-based semantic segmentation methods on SUN-RGB-D~\cite{song2015sun} and NYUDv2~\cite{silberman2012indoor} datasets using mean accuracy (mA), overall accuracy (OA), and mean intersection over union (mIoU) as the evaluation metrics. These results of various methods are taken from the original articles and shown in Table~\ref{table7}.

\begin{table}[t]
\centering
\caption{Evaluation performance regarding for RGB-D semantic segmentation methods on the SUN-RGB-D and NYUDv2. Note that the '\%' after the value is omitted and the symbol '--' means the results are unavailable.}
\label{table7}
\scriptsize
\renewcommand\arraystretch{1.4}
\centering
\setlength{\tabcolsep}{4mm}{
\begin{tabular*}{\columnwidth}{l|cc|cc}
\specialrule{1pt}{0pt}{0pt} 
\multirow{2}{*}{\textbf{Methods}}    & \multicolumn{2}{c|}{\textbf{NYUDv2}} & \multicolumn{2}{c}{\textbf{SUN-RGB-D}} \\ \cline{2-5}
& mAcc         & mIoU        & mAcc          & mIoU          \\\hline
Guo~et~al.~\cite{guo2018semantic}        & 46.3         & 34.8        & 45.7          & 33.7          \\
Wang~et~al.~\cite{wang2015towards}      & --           & 44.2        & --            & --            \\
Mousavian~et~al.~\cite{mousavian2016joint}  & 52.3         & 39.2        & --            & --            \\
Liu~et~al.~\cite{liu2018collaborative}          & 50.8         & 39.8        & 50.0          & 39.4          \\
Gupta~et~al.~\cite{guptalearning}        & 35.1         & 28.6        & --            & --            \\
Liu~et~al.~\cite{liu2018rgb}         & 51.7         & 41.2        & --            & --            \\
hazirbas~et~al.~\cite{hazirbas2016fusenet}  & --           & --          & 48.3          & 37.3          \\
Lin~et~al.~\cite{lin2017cascaded}          & --           & 47.7        & --            & 48.1          \\
Jiang~et~al.~\cite{jiang2017incorporating}      & --           & --          & 50.6          & 39.3          \\
Wang~et~al.~\cite{wang2018depth}         & 47.3         & --          & --            & --            \\
Cheng~et~al.~\cite{cheng2017locality}      & 60.7         & 45.9        & 58.0          & --            \\
Fan~et~al.~\cite{fan2017rgb}          & 50.2         & --          & --            & --            \\
Li~et~al.~\cite{li2016lstm}           & 49.4         & --          & 48.1          & --            \\
Qi~et~al.~\cite{qi20173d}           & 55.7         & 43.1        & 57.0          & 45.9          \\
Wang~et~al.~\cite{wang2016learning}         & 60.6         & 38.3        & 50.1          & 33.5         \\ \specialrule{1pt}{0pt}{0pt} 
\end{tabular*}}
\end{table}

We report the results of projected images, voxel, point clouds, and other representation semantic segmentation methods on S3DIS~\cite{armeni20163d} (both Area 5 and 6-fold cross-validation), ScanNet~\cite{dai2017scannet} (test sets), Semantic3D~\cite{hackel2017semantic3d} (reduced-8 subsets), and SemanticKITTI~\cite{behley2019semantickitti} (only xyz without RGB). We use mA, OA, and mIoU as the evaluation metrics. These results of various methods are taken from the original papers. Tables~\ref{table8} reports the results.

Point cloud semantic segmentation architectures typically focus on five main components: basic framework, neighborhood search, features abstraction, coarsening, and pre-processing. Below, we provide a more detailed discussion of each component.

\begin{table*}[t]
\centering
\caption{Evaluation performance regarding for projected images, voxel, point clouds and other representation semantic segmentation methods on the S3DIS, ScanNet, Semantic3D and SemanticKITTI. Note: the ‘\%’ after the value is omitted, the symbol ‘--’ means the results are unavailable, the dotted line means the subdivision of methods  according to the type of architecture.}
\label{table8}
\scriptsize

\centering
\renewcommand\arraystretch{1.05}
\setlength{\tabcolsep}{3.8mm}{\begin{tabular*}{0.92 \textwidth}{@{}l|c|ccc|cc|cc|cc}
\specialrule{1pt}{0pt}{0pt} 
\textbf{\multirow{3}{*}{Method}}    & \textbf{\multirow{3}{*}{Type}}             & \multicolumn{3}{c|}{\textbf{S3DIS}}           & \multicolumn{2}{c|}{\textbf{ScanNet}}  & \multicolumn{2}{c|}{\textbf{Semantic3D}} & \multicolumn{2}{c}{\textbf{SemanticKITTI}} \\
                            &                                   & \multicolumn{2}{c}{Area5} & 6-fold  & \multicolumn{2}{c|}{test set} & \multicolumn{2}{c|}{reduced-8}  & \multicolumn{2}{c}{only xyz}      \\
                            &                                   & mAcc        & mIoU        & mIoU    & oAcc          & mIoU         & oAcc          & mIoU           & mAcc             & mIoU           \\ \hline
Lawin et al.  \cite{lawin2017deep}        & \multirow{7}{*}{\begin{sideways}projection\end{sideways}} & --          & --          & --      & --            & --           & 88.9          & 58.5           & --               & --             \\
Boulch et al.  \cite{boulch2017unstructured}    &                                   & --          & --          & --      & --            & --           & 91.0          & 67.4           & --               & --             \\
Wu et al.  \cite{wu2018squeezeseg}         &                                   & --          & --          & --      & --            & --           & --            & --             & --               & 37.2           \\
Wang et al.  \cite{wang2018pointseg}          &                                   & --          & --          & --      & --            & --           & --            & --             & --               & 39.8           \\
Wu et al.  \cite{wu2019squeezesegv2}           &                                   & --          & --          & --      & --            & --           & --            & --             & --               & 44.9           \\
Milioto et al.  \cite{milioto2019rangenet++}    &                                   & --          & --          & --      & --            & --           & --            & --             & --               & 52.2           \\
Xu et al.  \cite{xu2020squeezesegv3}          &                                   & --          & --          & --      & --            & --           & --            & --             & --               & 55.9           \\

RangViT \cite{ando2023rangevit}          &                                   & --          & --          & --      & --            & --           & --            & --             & --               & 55.9           \\

RangFormer \cite{kong2023rethinking}          &                                   & --          & --          & --      & --            & --           & --            & --             & --               & 64.0           \\\hline

Tchapmi et al.  \cite{tchapmi2017segcloud}   & \multirow{3}{*}{\begin{sideways}voxel\end{sideways}}            & 57.35       & 48.92       & 48.92   & --            & --           & 88.1          & 61.30          & --               & --             \\
Meng et al.  \cite{meng2019vv}         &                                   & --          & 78.22       & --      & --            & --           & --            & --             & --               & --             \\
Liu et al.  \cite{liu20173DCNN}          &                                   & --          & 70.76       & --      & --            & --           & --            & --             & --               & --             \\\hline
PointNet  \cite{qi2017pointnet}            & \multirow{36}{*}{\begin{sideways}point\end{sideways}}    & 48.98       & 41.09       & 47.71   & --            & 14.69        & --            & --             & 29.9             & 17.9           \\
G+RCU  \cite{engelmann2018know}     &                                   & 59.10       &52.17       & 58.27   & 75.53         & --           & --            & --             & 57.59            & 29.9           \\
ESC  \cite{engelmann2017exploring}                &                                   & 54.06       &45.14       & 49.7    & 63.4          & --           & --            & --             & 40.9             & 26.4           \\
HRNN  \cite{ye20183d}                &                                   & 71.3        &53.4        & --      & 76.5          & --           & --            & --             & 49.2             & 34.5           \\
PointNet++  \cite{qi2017pointnet++}          &                                   & --          &50.04       & 54.4    & 71.40         &34.26        & --            & --             & --               & --             \\
PointWeb  \cite{zhao2019pointweb}            &                                   & 66.64       & 60.28       & 66.7    & 85.9          & --           & --            & --             & --               & --             \\
PointSIFT  \cite{jiang2018pointsift}           &                                   & --          &70.23       & 70.2    & --            &41.5         & --            & --             & --               & --             \\
Resurf  \cite{ran2022surface}           &                                   & 76.0          &68.9       & 74.3    & --            &70.0         & --            & --             & --               & --             \\

PointNeXt  \cite{qian2022pointnext}           &                                   & --          &70.5       & 74.9    & --            &71.2         & --            & --             & --               & --             \\

\textcolor{black}{PointVector  \cite{deng2023pointvector}}              &                                   & \textcolor{black}{78.1}          &\textcolor{black}{72.3}        & \textcolor{black}{78.4}    & \textcolor{black}{--}            &\textcolor{black}{--}         &\textcolor{black}{--}          &\textcolor{black}{ -- }          &\textcolor{black}{--}              & \textcolor{black}{--}             \\

\cline{3-11}

RSNet  \cite{huang2018recurrent}               &                                   & 59.42       &56.5        & 56.47   & --            & 39.35        & --            & --             & --               & --             \\
DPC  \cite{engelmann2020dilated}                 &                                   & 68.38       & 61.28       & --      & --            &59.2         & --            & --             & --               & --             \\
PointwiseCNN  \cite{hua2018pointwise}        &                                   & 56.5        &--          & --      & --            & --           & --            & --             & --               & --             \\
PCCN  \cite{wang2018deep}                &                                   & 67.01       & 58.27       & --      & --            & 49.8         & --            & --             & --               & --             \\

PointCNN  \cite{li2018pointcnn}           &                                   & 63.86       & 57.26       & 65.3    & 85.1          &45.8         & --            & --             & --               & --             \\
KPConv  \cite{thomas2024kpconvx}              &                                   & --          &67.1        & 70.6    & --            &66.6         & 92.9          & 74.6           & --               & --             \\

\textcolor{black}{KPConvX  \cite{thomas2019kpconv}}              &                                   & \textcolor{black}{78.7}          &\textcolor{black}{73.5}        & \textcolor{black}{--}    & \textcolor{black}{--}            &\textcolor{black}{76.3}         &\textcolor{black}{--}          &\textcolor{black}{-- }          &\textcolor{black}{--}              & \textcolor{black}{--}             \\

PointConv  \cite{wu2019pointconv}          &                                   & --          & 50.34       & --      & --            & 55.6         & --            & --             & --               & --             \\
A-CNN  \cite{komarichev2019cnn}               &                                   & --          & --          & --      & 85.4          & --           & --            & --             & --               & --             \\
RandLA-Net  \cite{hu2020randla}         &                                   & --          &--          & 70.0    & --            & --           & 94.8          & 77.4           & --               & 53.9           \\
PolarNet  \cite{zhang2020polarnet}            &                                   & --          & --          & --      & --            & --           & --            & --             & --               & 54.3           \\\cline{3-11}
DGCNN  \cite{wang2019dynamic}                &                                   & --          & 56.1        & 56.1    & --            & --           & --            & --             & --               & --             \\
SPG  \cite{landrieu2018large}                &                                   & 66.50       & 58.04       & 62.1    & --            & --           & 94.0          & 73.2           & --               & --             \\
SPH3D-GCN  \cite{lei2020spherical}           &                                   & 65.9        & 59.5        & 68.9    & --            &61.0         & --            & --             & --               & --             \\
DeepGCNs  \cite{li2019deepgcns}            &                                   & --          & 60.0        & --      & --            & --           & --            & --             & --               & --             \\
PointGCRNet  \cite{ma2020global}         &                                   & --          & 52.43       & --      & --            &60.8         & --            & --             & --               & --             \\
AGCN  \cite{xie2020point}                &                                   & --          & --          & 56.63   & --            & --           & --            & --             & --               & --             \\
PAN  \cite{feng2020point}                 &                                   & --          & 66.3        & --      & 86.7          & 42.1         & --            & --             & --               & --             \\
TGNet  \cite{li2019tgnet}               &                                   & --          & 58.7        & --      & 66.2          & --           & --            & --             & --               & --             \\
HDGCN  \cite{liang2019hierarchical}             &                                   & 65.81       &59.33       & 66.85   & --            & --           & --            & --             & --               & --             \\
3DContextNet  \cite{zeng20183dcontextnet}        &                                   & 74.5        &55.6        & 55.6    & --            & --           & --            & --             & --               & --             \\\cline{3-11}

PGCRNet  \cite{ma2020global}        &                                   &--        &54.4        & --    & --            & --           & --            &69.5              & --               & --             \\

AGCN  \cite{xie2020point}        &                                   & 74.5        &55.6        & 55.6    & --            & --           & --            & --             & --               & --             \\
PointANSL  \cite{yan2020pointasnl}        &                                   & --        &62.6        & 68.7    & --            & --           & 66.6            & --             & --               & --             \\

Point Transformer  \cite{zhao2021point}        &                                   & 76.5        &70.4        & 73.5    & --            & --           & --            & --             & --               & --             \\

Point Transformer v2  \cite{wu2022point}        &                                   & 77.9        &71.6        & --    & --            & 75.2           & --            & --             & --               & --             \\

\textcolor{black}{FPTransformer  \cite{he2024full}}              &                                   & \textcolor{black}{78.8}          &\textcolor{black}{73.1}        & \textcolor{black}{76.8}    & \textcolor{black}{--}            &\textcolor{black}{75.5}         &\textcolor{black}{--}          &\textcolor{black}{ -- }          &\textcolor{black}{--}              & \textcolor{black}{--}             \\

PatchFormer  \cite{zhang2022patchformer}        &                                   & --        &68.1        & --    & --            & --           & --            & --             & --               & --             \\
Fast Point Transformer  \cite{park2022fast}        &                                   & 77.3        &70.1        & --    & --            & --           & --            & --             & --               & --             \\
Stratify Transformer  \cite{lai2022stratified}        &                                   & 78.1        &72.0        & --    & --            & 73.7           & --            & --             & --               & --             \\

SphereFormer \cite{lai2023spherical}        &                                   & --        &--        &--    & --            & --           & --            & --             & --               & 78.4            \\

\textcolor{black}{ConDaFormer  \cite{duan2024condaformer}}              &                                   & \textcolor{black}{78.9}          &\textcolor{black}{73.5}        & \textcolor{black}{--}    & \textcolor{black}{--}            &\textcolor{black}{75.5}         &\textcolor{black}{--}          &\textcolor{black}{ -- }          &\textcolor{black}{--}              & \textcolor{black}{--}             \\

\textcolor{black}{Point Transformer v3  \cite{wu2024point}}              &                                   & \textcolor{black}{80.1}          &\textcolor{black}{74.3}        & \textcolor{black}{80.8}    & \textcolor{black}{78.6}            &\textcolor{black}{79.4}         &\textcolor{black}{--}          &\textcolor{black}{ -- }          &\textcolor{black}{--}              & \textcolor{black}{75.5}             \\
\hline

TangentConv  \cite{tatarchenko2018tangent}       & \multirow{6}{*}{\begin{sideways}others\end{sideways}}           & 62.2        & 52.8        & --      & 80.1          & 40.9         & 89.3          & 66.4           & --               & --             \\
SPLATNet  \cite{su2018splatnet}           &                                   & --          & --          & --      & --            & 39.3         & --            & --             & --               & --             \\
LatticeNet  \cite{rosu2019latticenet}         &                                   & --          & --          & --      & --            &64.0         & --            & --             & --               & 52.9           \\
Hung et al.  \cite{chiang2019unified}     &                                   & --          & --          & --      & --            & 63.4         & --            & --             & --               & --             \\
PVCNN  \cite{liu2019point}             &                                   & 87.12       & 58.98       & --      & --            & --           & --            & --             & --               & --             \\
MVPNet  \cite{jaritz2019multi}             &                                   & --          & --          & --      & --            &66.4         & --            & --             & --               & --             \\
BPNet  \cite{hu2021bidirectional}             &                                   & --          & --          & --      & --            &74.9         & --            & --             & --               & --             \\
\specialrule{1pt}{0pt}{0pt} 
\end{tabular*}}
\end{table*}

\vspace{1mm}\noindent\textbf{Basic framework:} is one of the main driving forces behind the development of 3D segmentation. Generally, two main basic frameworks exist, including PointNet and PointNet++. The PointNet framework utilizes shared MLPs to capture point-wise features and employs max-pooling to aggregate these features into a global representation. However, it cannot learn local features due to the absence of a defined local neighborhood. Additionally, the fixed resolution of the feature map makes it challenging to adapt to deep architectures. In contrast, the PointNet++ framework introduces a novel hierarchical learning architecture. It hierarchically defines local regions and progressively extracts features from these regions. This approach enables the network to capture local and global information, improving performance. As a result, many current networks adopt the PointNet++ framework or similar variations (such as 3D U-Net). This framework significantly reduces computational and memory complexities, particularly in high-level tasks like semantic segmentation, instance segmentation, and detection.

\vspace{1mm}\noindent\textbf{Neighborhood search:} To exploit the local features of point clouds, neighborhood point search is introduced into networks, including the KNN~\cite{zhao2021point},~\cite{ran2022surface},~\cite{qian2022pointnext}, ball search~\cite{hermosilla2018monte},~\cite{thomas2019kpconv},~\cite{lei2020spherical}, grid-based search~\cite{hua2018pointwise},~\cite{wu2022point} and tree-based search~\cite{lei2019octree}. KNN search retrieves the K closest neighbors to a query point based on a distance metric and hence lacks robustness to point clouds with varying densities. Some works integrate the dilated mechanism with the neighbor search to expand the receptive field~\cite{komarichev2019cnn},~\cite{li2018pointcnn},~\cite{li2019deepgcns}. Ball search involves finding all points within a specified radius (ball) around a query point. Similarly, grid-based search divides the point cloud space into a regular grid structure. Ball and grid-based algorithms are helpful for effectively capturing local structures and neighborhoods of varying densities.

\vspace{1mm}\noindent\textbf{Features abstraction:} In feature abstraction, commonly used methods include MLP-based, convolution-based, and transformer-based approaches. MLP often extracts features from individual points in point cloud data. MLP learns nonlinear point-level feature representations by bypassing each point's feature vectors through multiple fully connected layers. MLP offers flexibility and scalability in point cloud processing. Convolution operations on point clouds typically involve aggregating (low-level) information from local points to capture local structures and contextual information. In contrast, transformer-based methods establish correlations between high-level point information through the attention mechanism, which is more helpful for high-level tasks such as point cloud segmentation.
 
The essence of MLP-based, convolution-based, and transformer-based methods is to learn the relationships between points and obtain robust weights. In the context of a similar baseline architecture, the more comprehensive the learned point cloud relationship in the feature abstraction process, the stronger the robustness of the model becomes. Recently, MLP-based methods, such as Resurf~\cite{ran2022surface} and PointNeXt~\cite{qian2022pointnext}, exhibit better accuracy and efficiency, encouraging researchers to re-examine and further explore the potential of MLP-based approaches.

\vspace{1mm}\noindent\textbf{Coarsening}, also known as downsampling or subsampling, involves reducing the number of points in the point cloud while preserving the essential structures and features. Coarsening techniques include \textit{random sampling}~\cite{hu2020randla}, \textit{farthest point sampling}~\cite{qi2017pointnet, qi2017pointnet++}, \textit{tree-based} methods~\cite{lei2019octree} and mesh-based decimation~\cite{lei2023mesh}. This step helps to reduce computational complexity and improve efficiency in subsequent stages of the segmentation process. Random sampling is simple and computationally efficient but may not select optimal points for maintaining local and global structures. This can potentially lead to information loss in feature-rich regions. FPS is widely used in networks to ensure a more even spatial distribution of the selected points and help preserve global structures. However, local structures can still be destroyed with the farthest point sampling. Tree-based methods leverage hierarchical tree structures, such as an octree, to partition the point cloud and perform coarsening. Mesh-based methods must first convert the point cloud to a mesh before it can be decimated. This adds computational overhead to the already expensive mesh decimation process. Moreover, creating a mesh from complex and sparse point clouds obtained from LiDAR sensors is not always possible~\cite{lei2023mesh}.

The above methods are hand-crafted or engineered techniques that do not directly involve learning parameters from the data, which determines the sub-sampling pattern based on predefined rules or heuristics without explicitly optimizing for the task. Therefore, some works propose learnable coarsening methods that integrate a learnable layer into the coarsening module, such as pooling~\cite{groh2018flex},~\cite{lai2023spherical},~\cite{wu2022point},~\cite{zhao2021point}, and attention mechanism~\cite{yan2020pointasnl}.

\vspace{1mm}\noindent\textbf{Pre-processing:} is an essential step in point cloud semantic segmentation that involves preparing and transforming the raw point cloud data before feeding it into the segmentation network. Pre-processing aims to enhance the data's quality, consistency, and suitability for the segmentation task. Some common pre-processing aspects of point cloud segmentation include data normalization, outlier removal, data augmentation, and point registration.

Point clouds often have varying scales, which can negatively affect the performance of segmentation networks. Data normalization involves scaling the point cloud data to a standard range or unit sphere to ensure consistent scales across different points. For example, the number of ShapNet object points is generally fixed at 4096. For the complexity scene, early works~\cite{xu2021paconv},~\cite{qi2017pointnet} divided raw point clouds into smaller ones (e.g., 4096 points, 1m$^3$ blocks) so that the processing does not require large memory. However, this strategy might break down the semantic continuity of the scene. Recent works~\cite{qian2022pointnext},~\cite{lai2023spherical},~\cite{wu2022point},~\cite{ zhao2021point},~\cite{lei2023mesh} input the complete scene into the network, but that requires more computational sources. Moreover, these works tend to downsample the point cloud in the pre-processing stage.


\begin{table*}
\centering
\renewcommand\arraystretch{2}
\vspace{2mm}
\caption{Evaluation performance regarding for 3D instance segmentation methods on the ScanNet. Note: the '\%' after the value is omitted.}
\label{table9}
\scriptsize
\centering
\renewcommand\arraystretch{1.2}
\setlength{\tabcolsep}{1.4mm}{
\begin{tabular*}{0.98\textwidth}{@{}lccccccccccccccccccc@{}}
\specialrule{1pt}{0pt}{0pt} 
\textbf{Methods}      & \textbf{mAP}                 &bath.             & bed                & book.           & cabi.             &chair               & count.             & curt.             &desk                & door                & other               &pict.             & refr.             &shower.          & sink                & sofa                & table               &toilet              &wind.              \\\hline

GSPN~\cite{yi2019gspn}         & 30.6      & 50.0      & 40.5     & 31.1       & 34.8        & 58.9      & 5.4        & 6.8          & 12.6        & 28.3           & 29.0        & 2.8           & 21.9          & 21.4         & 33.1        & 39.6        & 27.5           & 82.1           & 24.5              \\

3D-SIS~\cite{hou20193d}      &38.2 &100 &43.2 &24.5 &19.0 &57.7 &1.3 &26.3 &3.3 &32.0 &24.0 &7.5 &42.2 &85.7 &11.7 &69.9 &27.1 &88.3 &23.5
 \\

3D-BoNet~\cite{yang2019learning}     &48.8 &100 &67.2 &59.0 &30.1 &48.4 &9.8 &62.0 &30.6 &34.1 &25.9 &12.5 &43.4 &79.6 &40.2 &49.9 &51.3 &90.9 &43.9 \\

SGPN~\cite{wang2018sgpn}         & 14.3    & 20.8      & 39.0      & 16.9        & 6.5         & 27.5      & 2.9      & 6.9           & 0           & 8.7            & 4.3           & 1.4         & 2.7             & 0            & 11.2        & 35.1       & 16.8           & 43.8           & 13.8                         \\

3D-MPA~\cite{engelmann20203d}       &61.1 &100 &83.3 &76.5 &52.6 &75.6 &13.6 &58.8 &47.0 &43.8 &43.2 &35.8 &65.0 &85.7 &42.9 &76.5 &55.7 &100 &43.0      \\

SoftGroup~\cite{vu2022softgroup}      &76.1 &100 &80.8 &84.5 &71.6 &86.2 &24.3 &82.4 &65.5 &62.0 &73.4 &69.9 &79.1 &98.1 &71.6 &84.4 &76.9 &100 &59.4   \\

SSTNet~\cite{liang2021instance}       &69.8 &100 &69.7 &88.8 &55.6 &80.3 &38.7 &62.6 &41.7 &55.6 &58.5 &70.2 &60.0 &100 &82.4 &72.0 &69.2 &100 &50.9  \\

\hline
3D-BEVIS~\cite{elich20193d}     & 24.8     & 66.7     & 56.6     & 7.6         & 3.5         & 39.4       & 2.7     & 3.5            & 9.8        & 9.8             & 3.0           & 2.5         & 9.8             & 37.5          & 12.6        & 60.4      & 18.1           & 85.4            & 17.1                   \\
PanopticFus.~\cite{narita2019panopticfusion} & 47.8    & 66.7      & 71.2     & 59.5        & 25.9        & 55.0        & 0      & 61.3           & 17.5        & 25.0           & 43.4          & 43.7       & 41.1             &85.7           & 48.5        & 59.1      & 26.7            &94.4            & 35.9                \\

OccuSeg~\cite{han2020occuseg}      &67.2 &100 &75.8 &68.2 &57.6 &84.2 &47.7 &50.4 &52.4 &56.7 &58.5 &45.1 &55.7 &100 &75.1 &79.7 &56.3 &100 &46.7\\

MTML~\cite{lahoud20193d}         &54.9 &100 &80.7 &58.8 &32.7 &64.7 &0.4 &81.5 &18.0 &41.8 &36.4 &18.2 &44.5 &100 &44.2 &68.8 &57.1 &100 &39.6 \\

PointGroup~\cite{jiang2020pointgroup} &63.6 &100 &76.5 &62.4 &50.5 &79.7 &11.6 &69.6 &38.4 &44.1 &55.9 &47.6 &59.6 &100 &66.6 &75.6 &55.6 &99.7 &51.3\\

HAIS~\cite{chen2021hierarchical}   &69.9 &100 &84.9 &82.0 &67.5 &80.8 &27.9 &75.7 &46.5 &51.7 &59.6 &55.9 &60.0 &100 &65.4 &76.7 &67.6 &99.4 &56.0\\

Dyco3D~\cite{he2021dyco3d}   &64.1 &100 &84.1 &89.3 &53.1 &80.2 &11.5 &58.8 &44.8 &43.8 &53.7 &43.0 &55.0 &85.7 &53.4 &76.4 &65.7 &98.7	&56.8
\\

DKNet~\cite{wu20223d}  &71.8 &100	&81.4 &78.2 &61.9 &87.2 &22.4 &75.1 &56.9 &67.7 	&58.5 &72.4 &63.3 &98.1 &51.5 &81.9 &73.6 &100 &61.7
 \\

ISBNet~\cite{ngo2023isbnet}   &76.3	&100 &87.3 	&71.7	&66.6 &85.8 &50.8 &66.7 &76.4 &64.3 &67.6 &68.8 &82.5 &100 &77.3 &74.1 &77.7 & 100	&55.6
 \\

\textcolor{black}{Spherical Mask~\cite{shin2024spherical}} &\textcolor{black}{81.2}  &\textcolor{black}{100}	&\textcolor{black}{97.3} 	&\textcolor{black}{85.2} 	&\textcolor{black}{71.8} 	&\textcolor{black}{91.7} 	&\textcolor{black}{57.4 }	&\textcolor{black}{67.7} 	&\textcolor{black}{74.8} 	&\textcolor{black}{72.9 }	&\textcolor{black}{71.5} 	&\textcolor{black}{79.5} 	&\textcolor{black}{80.9}	&\textcolor{black}{100} 	&\textcolor{black}{83.1} 	&\textcolor{black}{85.4} 	&\textcolor{black}{78.7} 	&\textcolor{black}{100} 	&\textcolor{black}{63.8} 
 \\

\textcolor{black}{SPFormer~\cite{sun2023superpoint}}   &\textcolor{black}{77.0} 	&\textcolor{black}{90.3} 	&\textcolor{black}{90.3} 	&\textcolor{black}{80.6} 	&\textcolor{black}{60.9} 	&\textcolor{black}{88.6} 	&\textcolor{black}{56.8} 	&\textcolor{black}{81.5} 	&\textcolor{black}{70.5} 	&\textcolor{black}{71.1} 	&\textcolor{black}{65.5} 	&\textcolor{black}{65.2} 	&\textcolor{black}{68.5} 	&\textcolor{black}{100} 	&\textcolor{black}{78.9} 	&\textcolor{black}{80.9} 	&\textcolor{black}{77.6} 	&\textcolor{black}{100} 	&\textcolor{black}{58.3} 
 \\

\textcolor{black}{Mask3D~\cite{schult2023mask3d}}   &\textcolor{black}{78.0}	&\textcolor{black}{100} 	&\textcolor{black}{78.6} 	&\textcolor{black}{71.6} 	&\textcolor{black}{69.6}	&\textcolor{black}{88.5} 	&\textcolor{black}{50.0} 	&\textcolor{black}{71.4} 	&\textcolor{black}{81.0} 	&\textcolor{black}{67.2}	&\textcolor{black}{71.5} 	&\textcolor{black}{67.9} 	&\textcolor{black}{80.9}	&\textcolor{black}{100} 	&\textcolor{black}{83.1}	&\textcolor{black}{83.3} 	&\textcolor{black}{78.7} 	&\textcolor{black}{100} 	&\textcolor{black}{60.2}
 \\

\textcolor{black}{QueryFormer~\cite{lu2023query}}   &\textcolor{black}{78.7} 	&\textcolor{black}{100} 	&\textcolor{black}{93.3} &\textcolor{black}{60.1} &\textcolor{black}{75.4} &\textcolor{black}{88.6} &\textcolor{black}{55.8} &\textcolor{black}{66.1} &\textcolor{black}{76.7} 	&\textcolor{black}{66.5} 	&\textcolor{black}{71.6} &\textcolor{black}{63.9} &\textcolor{black}{80.8} 	&\textcolor{black}{100} 	&\textcolor{black}{84.4} 	&\textcolor{black}{89.7} 	&\textcolor{black}{80.4} 	&\textcolor{black}{100} 	&\textcolor{black}{62.4}
 \\

\textcolor{black}{OneFormer~\cite{kolodiazhnyi2024oneformer3d}}   &\textcolor{black}{80.1} &\textcolor{black}{100} 	&\textcolor{black}{97.3 }	&\textcolor{black}{90.9} &\textcolor{black}{69.8} &\textcolor{black}{92.8} &\textcolor{black}{58.2} &\textcolor{black}{66.8} &\textcolor{black}{68.5} &\textcolor{black}{78.0} &\textcolor{black}{68.7} &\textcolor{black}{69.8} &\textcolor{black}{70.2} 	&\textcolor{black}{100} &\textcolor{black}{79.4} &\textcolor{black}{90.0} &\textcolor{black}{78.4} &\textcolor{black}{98.6} &\textcolor{black}{63.5}
 \\

\specialrule{1pt}{0pt}{0pt} 
\end{tabular*}
}
\vspace{4mm}
\end{table*}

\subsection{Results for 3D Instance Segmentation}
\begin{figure}[t]
\centering
\includegraphics[width=\columnwidth]{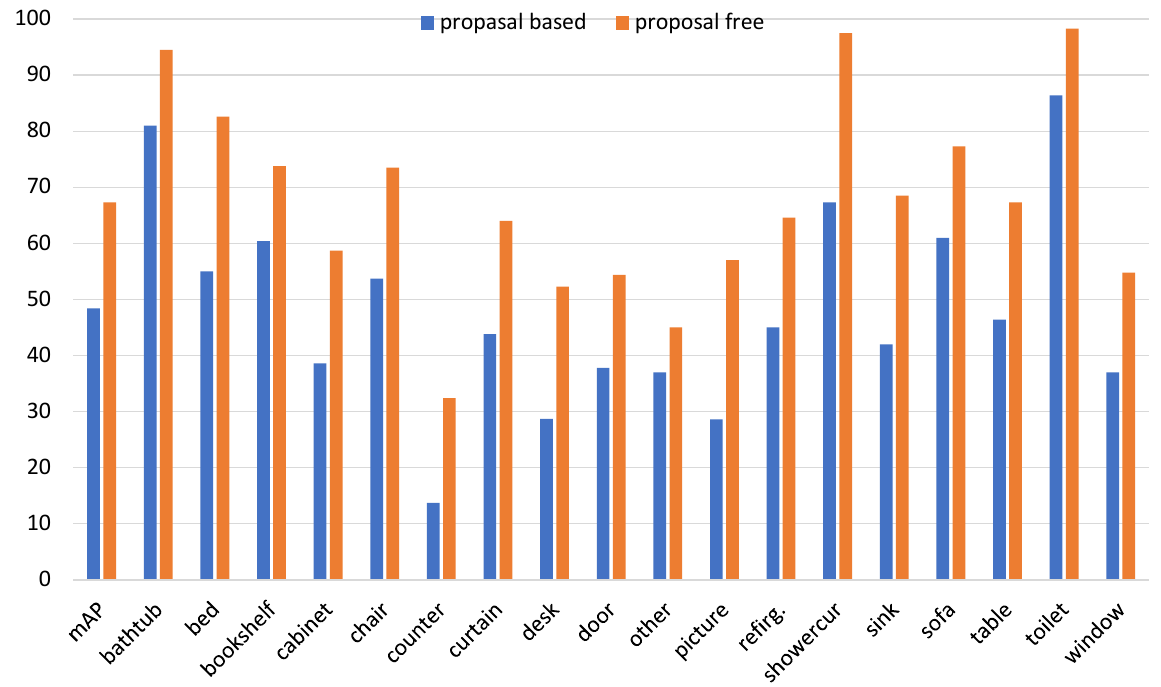}
\vspace{-2mm}
\caption{\textcolor{black}{Evaluation performance regarding for 3D instance segmentation architecture, including proposal based and proposal free, on the different class of ScanNet. For simplicity, we omit the '\%' after the value.}}
\label{evalutiononscannet}
\end{figure}

We report the results of 3D instance segmentation methods on ScanNet~\cite{dai2017scannet} datasets and choose mAP as the evaluation metrics. The results of these methods are taken from the ScanNet Benchmark Challenge website, shown in Table \ref{table9} and summarized in Figure~\ref{evalutiononscannet}. \textcolor{black}{The Spherical Mask~\cite{shin2024spherical} has state-of-the-art performance, with 81.2\% average precision on the ScanNet dataset at this view.} It also achieves the best instance segmentation performance in most classes, including `bathtub,' `shower~curtain,' `toilet,' and so on.

Most methods have better segmentation performance on large-scale classes such as `bathtub' and `toilet' and have poor segmentation performance on small-scale classes such as `counter,' `desk,' and `picture.' Therefore, the instance segmentation of small objects is a prominent challenge.

In proposal-based methods, specifically the 2D embedding propagating-based methods such as 3D-BEVIS~\cite{elich20193d} and PanoticFusion~\cite{narita2019panopticfusion}, they tend to exhibit poorer performance compared to other proposal-free methods. This is primarily because simple embedding propagation techniques are more susceptible to error labels, leading to inaccuracies in the instance segmentation results.

Proposal-free methods demonstrate superior performance than proposal-based methods in instance segmentation across all classes, particularly for small objects like `curtains,' `pictures,' `shower~curtains,' and `sinks.' Unlike proposal-based methods that rely on the accuracy of proposal generation, proposal-free methods circumvent this issue entirely. They directly consider the entire point cloud and its global features, enabling more precise and comprehensive instance segmentation. By avoiding the need for proposal generation, proposal-free methods can achieve better results by considering the overall context and characteristics of the point cloud.

\begin{table}[tbp]
\centering
\caption{Evaluation performance regarding for 3D part segmentation on the ShapeNet. Note: the '\%' after the value is omitted, the symbol '--' means the results are unavailable.}
\vspace{2mm}
\label{table10}
\scriptsize
\renewcommand\arraystretch{1.42}
\setlength{\tabcolsep}{2mm}{
\begin{tabular*}{0.95\columnwidth}{l|c|l|c}
\specialrule{1pt}{0pt}{0pt} 
\textbf{Methods}              &\textbf{ Ins.   mIoU} & \textbf{Methods}                 & \textbf{Ins.   mIoU} \\\hline
VV-Net~\cite{meng2019vv}      & 87.4        & LatticeNet~\cite{su2018splatnet}     &83.9       \\
SSCNet~\cite{graham20183d}       & 86.0        & SGPN~\cite{wang2018sgpn}            & 85.8        \\
PointNet~\cite{qi2017pointnet}     & 83.7        & ShapePFCN~\cite{kalogerakis20173d}      &  {88.4}        \\
PointNet++~\cite{qi2017pointnet++}   & 85.1        & VoxSegNet~\cite{wang2019voxsegnet}      & 87.5        \\
3DContextNet~\cite{zeng20183dcontextnet} & 84.3        & Pointgrid\cite{le2018pointgrid}      & 86.4        \\
RSNet~\cite{huang2018recurrent}        & 84.9   &KPConv~\cite{thomas2019kpconv}       & 86.4            \\
MCC~\cite{hermosilla2018monte}          & 85.9        & SO-Net~\cite{li2018so}        & 84.9        \\
PointConv~\cite{wu2019pointconv}    & 85.7        & PartNet~\cite{yu2019partnet}        & 87.4        \\
       
DGCNN~\cite{wang2019dynamic}         & 85.1        & SyncSpecCNN~\cite{yi2017syncspeccnn}    & 84.7        \\
SPH3D-GCN~\cite{lei2020spherical}    & 86.8        & KCNet~\cite{yi2017syncspeccnn}         & 84.7        \\
AGCN~\cite{xie2020point}        & 85.4  &PointCNN~\cite{li2018pointcnn}     & 86.1                 \\
PCCN~\cite{wang2018deep}         & 85.9        & SpiderCNN~\cite{xu2018spidercnn}      & 85.3        \\
Flex-Conv~\cite{groh2018flex}    & 85.0        & FeaStNet~\cite{verma2018feastnet}       & 81.5        \\
$\psi$-CNN~\cite{lei2019octree}        & 86.8        & Kd-Net~\cite{klokov2017escape}         & 82.3        \\
SPLATNet~\cite{su2018splatnet}    & 84.6        & O-CNN~\cite{wang2017cnn}         & 85.9  \\ 
DRGCNN~\cite{yue2022drgcnn} & 86.2 & \textcolor{black}{PointVector~\cite{deng2023pointvector}} &\textcolor{black}{86.9 }    \\

 \textcolor{black}{SPoTr~\cite{park2023self}} &\textcolor{black}{87.2 }  & \textcolor{black}{PointNext~\cite{qian2022pointnext}} &\textcolor{black}{87.1 }    \\

\textcolor{black}{CurveNet~\cite{xiang2021walk}} &\textcolor{black}{86.8 }  & \textcolor{black}{PointMLP~\cite{marethinking}} &\textcolor{black}{86.1 }    \\

\textcolor{black}{Point Transformer~\cite{zhao2021point}} &\textcolor{black}{86.6} &\\
\specialrule{1pt}{0pt}{0pt} 
\end{tabular*}
}
\end{table}

\subsection {Results for 3D Part Segmentation}
We report the results of 3D part segmentation methods on ShapeNet~\cite{yi2016scalable} datasets and use Ins.~mIoU as the evaluation metric. These results of various techniques are taken from the original papers and shown in Table \ref{table10}. We can find that the part segmentation performance of all methods is quite similar. One underlying assumption is that objects in ShapeNet datasets are synthetic, normalized in scale, aligned in pose, and lack scene context. This makes it difficult for part segmentation networks to extract rich context features. Another underlying assumption is that the point clouds in the synthetic scenes without background noise are simpler and cleaner than the ones in the real scenes, so the geometric features of point clouds are easy to exploit. The accuracy performance of various part segmentation networks is challenging to distinguish effectively.

\section{\textcolor{black}{Challenges and Discussions}}\label{section8}
\textcolor{black}{3D segmentation using deep learning techniques has made significant progress in recent years. However, this is just the beginning, and considerable challenges lie ahead. Below, we present some challenges and identify potential research directions.}

\vspace{2mm}
\noindent \textcolor{black}{\textbf{Data Scarcity and Quality:} Annotated 3D datasets are less abundant than 2D image datasets, This is because creating high-quality 3D annotations requires significant time and expertise, limiting the availability of large-scale datasets for training deep learning models. 3D data from various sensors often contain noise, missing regions, and inconsistencies. These issues pose significant challenges for accurate segmentation. Below, we list some promising research directions aimed at addressing this challenge.}

\begin{itemize}
\item \textcolor{black}{\textit{Synthetic data generation} gradually plays an important role in 3D segmentation due to the low cost and diverse scenes that can be generated~\cite{brodeur2017home},~\cite{wu2018building} compared to real dataset. Besides, Synthetic data can  include detailed important semantic information, such as material and texture information, which is essential for segmentation with similar color or geometric information.} 

\item \textcolor{black}{\textit{Transfer Learning} involves leveraging pre-trained models to improve performance on specific tasks, where the models are generally trained on large, often unrelated datasets~\cite{wu2023sim2real},~\cite{zhou2024dynamic}. In the context of 3D segmentation, transfer learning can significantly enhance accuracy performance and address challenges related to data scarcity and quality.}

\item \textcolor{black}{\textit{Weakly-Supervised Learning} minimizes the need for extensive and precise annotations by using less detailed supervision. This approach effectively reduces annotation effort while still improving model training.  It addresses data scarcity by making use of weaker or less accurate labels~\cite{su2023weakly},~\cite{ shi2022weakly},~\cite{kweon2024weakly}, and it helps manage data quality by being resilient to imperfections in the provided labels.}

\item \textcolor{black}{\textit{Semi-Supervised Learning} combines labeled and unlabeled data to enhance model performance and lower annotation costs~\cite{li2024density}. By leveraging a small amount of labeled data along with a larger pool of unlabeled data, this method improves generalization and robustness. It is particularly effective in situations where labeled data is limited but a larger volume of unlabeled data is available.}

\item \textcolor{black}{\textit{Unsupervised Learning} operates without labeled data, focusing on discovering hidden structures and patterns within the data~\cite{zhang2023growsp},~\cite{xie2020pointcontrast}. This approach is highly adaptable and resilient to data quality issues, making it ideal for scenarios with limited or noisy data~\cite{xiao2023unsupervised}. It learns valuable representations and insights from the data itself, without relying on annotated labels.}
\end{itemize}

\vspace{2mm}
\noindent \textcolor{black}{\textbf{Generalization and Robustness:} Models trained on specific datasets may not generalize well to different types of 3D data, such as those acquired from different sensors or environments. Developing robust models that perform well across various domains remains a challenge. Besides, Variations in object shapes, sizes, and orientations in real-world  can affect model performance. Ensuring robustness to such variations is critical for reliable 3D segmentation. Below
we list promising research directions for improving the generalization and robustness of point cloud processing.
}

\begin{itemize}
\item \textcolor{black}{\textit{Domain Adaptation} is a technique used to adapt models so they can perform effectively across different domains or environments, especially when there is a shift or discrepancy between the source domain and the target domain\textcolor{black}{~\cite{zhao2021epointda},~\cite{yi2021complete}}. Domain adaptation technique, including aligning features, reweighing instances, fine-tuning and self-training, is especially important in scenarios where labeled data in the target domain is limited or unavailable.}

\item \textcolor{black}{\textit{Multi-Modal Approaches} use data from multiple sensor modalities (e.g., LiDAR, image, voice, language) to enhance robustness and generalization. By integrating diverse sources of information, these approaches provide a more comprehensive understanding of the environment~\cite{hu2021bidirectional}\textcolor{black}{,~\cite{li2023mseg3d}}. This integration helps mitigate the limitations of individual modalities, offering richer feature representations and better handling of variability and uncertainty. }

\item \textcolor{black}{\textit{Large Model Approaches} often have deeper architectures and more parameters, allowing them to capture complex features and patterns. Popular approaches can include segmenting point clouds with large image models (such as SAM~\cite{kirillov2023segment},~\cite{wei2024semantic}) and natural language models like ChatGPT. The advanced capabilities of large models enable them to capture intricate patterns and semantic relationships, leading to improved performance and accuracy in segmentation tasks.}

\item \textcolor{black}{\textit{Meta-Learning} trains models to adapt rapidly to new, unseen data with minimal additional training\textcolor{black}{~\cite{zhao2021few}}. By optimizing the learning process and leveraging prior knowledge from various tasks, meta-learning enhances a model's ability to generalize, perform well with few examples, and transfer knowledge across different tasks.}

\end{itemize}

\vspace{2mm}
\noindent \textcolor{black}{\textbf{Computational Complexity:} Processing 3D data is computationally intensive due to its high dimensionality and volume. Training deep learning models on 3D datasets requires substantial computational resources, including powerful GPUs and extensive memory. As the size and resolution of 3D data increase, scalability becomes a concern. Efficiently managing memory and computational resources is essential for handling large-scale 3D datasets. Below, we list promising research directions for improving the computational efficiency of models.}

\begin{itemize}
\item \textcolor{black}{\textit{Efficient Architectures} aim to balance computational efficiency with high performance. Techniques such as sparse representations learning\textcolor{black}{~\cite{razani2021gp}}, and neural architecture search\textcolor{black}{~\cite{hu2020randla}} help achieve this balance by reducing resource requirements while maintaining effective segmentation capabilities. These optimizations are crucial for deploying 3D segmentation models in scenarios.}

\item \textcolor{black}{\textit{Parallel Computing} leverages multiple processing units to handle the large-scale computations required for segmenting 3D data efficiently. Techniques such as multi-core CPUs, GPUs, distributed computing, and parallel algorithms help improve processing speed, scalability, and efficiency. This approach is crucial for applications that require real-time processing and accurate segmentation of complex 3D datasets.}

\item \textcolor{black}{\textit{Model Compression} involves techniques such as pruning, quantization, knowledge distillation\textcolor{black}{~\cite{hou2022point},~\cite{ji2022structural}}, low-rank factorization, efficient network design, and model sharing to reduce model size and computational requirements while preserving performance. These methods help make 3D segmentation models more practical and efficient for deployment in resource-constrained environments and applications that require real-time processing.}

\end{itemize}

\vspace{2mm}
\noindent \textcolor{black}{\textbf{Interpretability and Explainability:} Deep learning models for 3D segmentation, particularly those with complex architectures, often lack interpretability\textcolor{black}{~\cite{atik2024explainable}}. Understanding the decision-making process of these models is crucial for gaining trust and ensuring reliable deployment in critical applications. Providing clear and understandable explanations for segmentation results is important, especially in fields such as medical imaging and cultural heritage preservation, where decisions must be transparent and justifiable. Some promising research directions for improving interpretability
and explainability of deep learning models are listed below.}

\begin{itemize}

\item \textcolor{black}{\textit{Explainable AI (XAI) Methods} include visualization techniques (saliency and activation maps), attention mechanisms, feature visualization, surrogate models, rule-based explanations, counterfactual explanations, and layer-wise relevance propagation. These methods enhance the interpretability and transparency of 3D segmentation models by providing insights into how decisions are made and which features are most influential. This facilitates better understanding, trust, and usability of AI systems across various applications.}

\item \textcolor{black}{\textit{Visualization and Post-Hoc Analysis} involve techniques such as saliency maps, activation maps, attention maps, feature visualization, surrogate models, rule-based explanations, counterfactual explanations. These methods help analyze and visualize model predictions and decision processes, offering insights into how 3D segmentation models work and enabling better understanding, debugging, and refinement of AI systems.}

\item \textcolor{black}{\textit{Interactive Tools} are specialized interfaces designed to help users visualize, explore, and understand the outputs of 3D segmentation models. Given the complexity of 3D data and the outputs of segmentation models, these tools are essential for gaining insights and making informed decisions about model performance and data interpretation.}

\end{itemize}

\section{Conclusion}\label{section9}
In conclusion, we present a comprehensive survey of recent developments in 3D segmentation utilizing deep learning techniques. Through this survey, we devolved into 3D semantic segmentation, 3D instance segmentation, and 3D part segmentation; we have covered around 180 methodologies, each with its strengths and weaknesses. Our performance comparison highlights the current state-of-the-art approaches, offering valuable insights into their efficacy across different datasets and application scenarios. However, it's important to note that the field is dynamic, and new methods are continually emerging, promising even greater accuracy, efficiency, and versatility advancements. Moreover, we have identified several promising research directions for further investigation. These include the exploration of novel architectures tailored specifically for 3D segmentation tasks, integrating multi-modal data sources to enhance segmentation accuracy, and the development of techniques for handling large-scale and dynamic 3D environments. By pushing the boundaries of 3D segmentation research, we can unlock new possibilities across a wide range of domains, from medical imaging and robotics to autonomous driving and virtual reality. Utilizing recent advances in generative AI, we can tackle the complexities of 3D data and advance 3D segmentation.

\printcredits

\bibliographystyle{cas-model2-names}

\bibliography{reference}

\end{document}